\RequirePackage{etoolbox}
\csdef{input@path}%
{%
 {sty/},
 {img/},
}
\documentclass{elsevierbook}

\usepackage{natbib}
\usepackage{amssymb}
\usepackage{amsthm}
\usepackage{amsthm}
\usepackage{amsmath,wasysym}
\usepackage{matlab-prettifier}
\usepackage{array}
\usepackage{tabularx}
\usepackage[export]{adjustbox}
\usepackage{hyperref}
\usepackage{amssymb}
\usepackage{mathrsfs}
\usepackage{algorithm}
\usepackage{relsize}
\usepackage{xcolor}
\usepackage{longtable}
\usepackage{hyperref}
\def \T{\mathsf{T}}
\def\H{\mathsf{H}}

\newcommand*{\E}[1]{\mathsf{E}\left\{ #1 \right\}}

\DeclareMathAlphabet\mathbfcal{OMS}{cmsy}{b}{n}

\def \0{\mathbf{0}}

\newcounter{example}
\newenvironment{example}[1][]{\refstepcounter{example}\par\medskip
   \noindent \textbf{Example~\theexample. #1} \rmfamily}{\medskip}

\begin{document}

\Frontmatter

\Mainmatter

%
%
%
\begin{frontmatter}

\setcounter{chapter}{5}
\renewcommand{\thechapter}{\arabic{chapter}}
\chapter{Hypercomplex Widely Linear Processing}\label{chap1}
\subchapter{Fundamentals for Quaternion Machine Learning}

\begin{aug}
\author[addressrefs={ad1}]%
  {\fnm{Sayed Pouria}   \snm{Talebi}}%
\author[addressrefs={ad2}]%
 {\fnm{Clive} \snm{Cheong Took}}%
\address[id=ad1]%
  {Computer Science Department, University of Roehampton, London WC2R-2LS, U.K.}
\address[id=ad2]%
 {Electronic~Engineering~Department, Royal Holloway University of London, TW20 0EX, U.K.}%
\end{aug}



\end{frontmatter}










\section*{Abstract}

Numerous attempts have been made to replicate the success of complex-valued algebra in engineering and science to other hypercomplex domains such as quaternions, tessarines, biquaternions, and octonions. Perhaps, none have matched the success of quaternions. The most useful feature of quaternions lies in their ability to model three-dimensional rotations which, in turn, have found various industrial applications such as in aeronautics and computer graphics. Recently, we have witnessed a renaissance of quaternions due to the rise of machine learning. To equip the reader to contribute to this emerging research area, this chapter lays down the foundation for

\begin{itemize}
\item augmented statistics for modelling quaternion-valued random processes,
\item widely linear models to exploit such advanced statistics,
\item quaternion calculus and algebra for algorithmic derivations,
\item mean square estimation for practical considerations.
\end{itemize}
For ease of exposure, several examples are offered to facilitate the learning, understanding, and (hopefully) the adoption of this multidimensional domain.

\vspace{0.32cm}

\noindent\textbf{\textit{Keywords}}: Quaternion Algebra; Augmented Quaternion Variable; Augmented Quaternion Statistics; $\mathbb{HR}$-Calculus; Quaternion-Valued Learning Techniques.  

\section*{Mathematical Nomenclature}

\noindent\textbf{\textit{Nomenclature}}: Scalars, column vectors, and matrices are denoted by lowercase, bold lowercase, and bold uppercase letters respectively. The remainder of nomenclature is summarised as follows:

\begin{longtable}{l l}
$\mathbb{N}$ & set of natural numbers
\\
$\mathbb{R}$ & set of real-valued numbers
\\
$\mathbb{R}^{+}$ & set of positive real-valued  numbers
\\
$\mathbb{C}$ & set of complex-valued numbers
\\
$\mathbb{H}$ & set of quaternion-valued numbers
\\
$\Re\{\cdot\}$ & operator returning the real component
\\
$\Im\{\cdot\}$ & operator returning the imaginary component
\\
$\{\imath,\jmath,\kappa\}$ &  imaginary units spanning the imaginary subspace of $\mathbb{H}$
\\
$\eta$ & a generic imaginary unit, which can be one of $\{\imath,\jmath,\kappa\}$ 
\\
$\Im_{\chi}\left(\cdot\right)$ & operator returning the imaginary component alongside $\chi\in\{\imath,\jmath,\kappa\}$
\\
$\|\cdot\|$ & second-order norm
\\
$\mathbf{I}$ & identity matrix of appropriate size
\\
$(\cdot)^{*}$ & conjugate operator
\\
$\left(\cdot\right)^{\T}$ & transpose operator
\\
$\left(\cdot\right)^{\H}$ & Hermitian operator
\\
$e$ & Euler's number
\\
$\E{\cdot}$ & statistical expectation operator
\\
$\partial$ & partial derivative operator
\\
$\mathsf{d}$ & total differential operators
\\
$\nabla_{\chi}$ & gradient operator with respect to $\chi$
\\
$\text{det}\left(\cdot\right)$ & determinant operator
\\
$\langle\cdot,\cdot\rangle$ & inner product
\\
$\times$ & cross product
\end{longtable}
\section{Introduction}
\indent Multidimensional signal processing is typically associated with the real domain, $\mathbb{R}$, and the vector space. Perhaps, a possible explanation for this assumption lies in the maturity of these domains over the hypercomplex domain of quaternions. Indeed, quaternions remain a relatively unexplored area in multidimensional signal processing. This presents opportunities for further research. To exploit such opportunities, this chapter aims to present the fundamentals of this hypercomplex domain based on the so-called widely linear processing model introduced in Section~\ref{Sec:MMSE}. This model implies that considering a quaternion variable in isolation is not sufficient to describe or capture the complete information of the quaternion. The aim of this chapter is to address this shortcoming. We first start with the history of quaternions to put in context the recent advances in quaternion-valued signal processing.  

The need for a number system beyond $\mathbb{R}$ became pressing in the $16^{\text{th}}$ century with the quest to derive a general solution to find roots of polynomials of arbitrary degree. Most notable in this area is the work of Niccolo Tartagila and Girolamo Cardano, who contributed to closed-form solutions for finding roots of polynomials of third and fourth degree~\cite{DB}. Rafael Bombelli, in his work on the roots of cubic polynomials introduced the symbol $\sqrt{-1}$ and showed that to solve for the roots of such polynomials it is necessary to perform calculations in $\mathbb{C}$~\cite{DB}. The timeline on the research into roots of polynomials culminates in the fundamental theory of algebra by Carl Friedrich Gauss, proving that polynomials of degree $n$ have exactly $n$ roots in $\mathbb{C}$. Perhaps the best insight to complex-valued numbers comes from Leonhard Euler, who expressed complex numbers in their polar format, that essentially equate multiplication and division to rotations in the two-dimensional plane of the complex field. Regarded as ``\textit{the most remarkable formula in mathematics}'' by Richard Feynman, the formulation introduced by Euler later became the cornerstone of complex-valued transforms of real-valued signals for frequency analysis. In all fields of engineering and physics, energy waves, e.g. sound, electromagnetic and gravitational waves, are transformed into complex-valued frequency domain representations where solving the differential equations that govern their behaviour is a more straightforward affair~\cite{Comms,Madar-1-2,Quantum-Physics}. 

Although complex-valued signals could be processed as a vector of their real and imaginary parts, this approach leads to the loss of important notions such as phase and frequency. The need for complex-valued adaptive processing that treated complex-valued signals directly in the complex domain was recognized by Bernard Widrow leading to the derivation of the complex least mean square algorithm, an extension of his famous least mean square algorithm~\cite{CLMS}. Perhaps one of the most insightful developments in this field is that the covariance, $\E{\mathbf{z}\mathbf{z}^{\H}}$, needs to be augmented with a pseudo-covariance $\E{\mathbf{z}\mathbf{z}^{\T}}$ to represent the full second-order statistical information~\cite{DB,SOACRV,CI}. To better demonstrate this point consider the standard covariance
\begin{equation}
\E{\mathbf{z}\mathbf{z}^{\H}}=\E{\mathbf{x}\mathbf{x}^{\T}}+\E{\mathbf{y}\mathbf{y}^{\T}}+\jmath
\left(\E{\mathbf{y}\mathbf{x}^{\T}}-\E{\mathbf{x}\mathbf{y}^{\T}}\right)
\label{eq:ComplexCV}
\end{equation}
and pseudo-covariance
\begin{equation}
\E{\mathbf{z}\mathbf{z}^{\T}}=\E{\mathbf{x}\mathbf{x}^{\T}}-\E{\mathbf{y}\mathbf{y}^{\T}}+\jmath\left(\E{\mathbf{x}\mathbf{y}^{\T}}+\E{\mathbf{y}\mathbf{x}^{\T}}\right)
\label{eq:ComplexPCV}
\end{equation}
where $\mathbf{z}=\mathbf{x}+\jmath\mathbf{y}$ with $\{\mathbf{x},\mathbf{y}\}\in\mathbb{R}^{n}$. Note that the second-order information of neither the real nor imaginary segments of $\mathbf{z}$ is inferable from the standard or pseudo-covariances. However, in conjunction, second-order information of the real and imaginary segments of $\mathbf{z}$ and their second-order dependencies can be inferred via linear manipulations of the covariance and pseudo-covariance. 

The quaternion story begins with Sir William Rowan Hamilton. Concerned with constructing a framework for modelling three-dimensional spaces in a manner analogous to that achieved through complex-valued algebra for two-dimensional spaces, Hamilton came to the conclusion that this can only be made possible through a number system that has three imaginary units that admit a special set of relations. As the story is told, these relations dawned on Hamilton during a walk with his wife. Recognizing their importance and not wanting to miss the opportunity, Hamilton carved these relations onto Brougham Bridge in the best known act of scientific graffiti~\cite{QuatBook}. Hamilton focused on deriving a model to explain movement and orientation in three-dimensional spaces, resulting in an algebra that is most effective in modelling real-world phenomena in our three-dimensional surroundings. Moreover, the resulting quaternion algebra itself is a four-dimensional one. Hamilton surmised that quaternions could model space and time simultaneously in a letter to Sir John Herschel~\cite{Letter}, writing; ``\textit{... how the one of time, of space the three, might in the chain of symbols, girdled be. It is not so much to be wondered that they should have let me to strike out some new lines of research, which former methods have failed to suggest.}'' Indeed, quaternions have seen use in quantum mechanics and formulation of space-time relativity where, as one example, the quaternion four-dimensional algebra has allowed to simultaneously formulate rotational and compression gradients~\cite{QQM,Quantum-Physics}.

The most common use of quaternions has been in modelling rotation and orientation, ranging from aerospace applications~\cite{Kuipers,Survey} to computer graphics~\cite{CG}. However, the ability of quaternions to model physical phenomena has been useful in a wider range of applications. One example is the work of James Clark Maxwell on the derivation of his famous equations of electromagnetism~\cite{MaxEqu,QuatBook}.\footnote{Despite Maxwell being one of the proponents for the use of quaternion algebra, his work on electromagnetism was later re-derived in their current vector algebra formulation by Oliver Heaviside~\cite{HeavisideIntro,Heaviside}.} Modern applications of the high-dimensional division algebra that quaternions provide include modelling and analysis of multi-phase power systems~\cite{3PhaseMe,Oleg,3B}, body motion tracking~\cite{QRKHS}, three-dimensional sound processing~\cite{3DS,MUSIC}, colour image processing~\cite{Image-2}, communication techniques that adopt space-time-polarisation multiplexing~\cite{Comm1,Comm2,CommME}, and quantum computing~\cite{QuantComplier,QML}. 

Akin to the importance of complex-valued adaptive processing techniques for complex-valued signals, it has come to the attention of the signal processing, machine learning, and control communities that it is best to process quaternion-valued signals directly in the quaternion domain where they reside. This reserves the physical meaning of the derived models, allows for straightforward analysis of the operations and obtained results, as well as, preserving the notion of division within the algebra used for deriving adaptive signal processing techniques. In order to make the quaternion domain accessible, this chapter aims to equip the reader with the necessary algebraic, statistical, and calculus tools to derive and analyse in this multidimensional domain. In  Section~\ref{Sec:QA}, quaternion algebra and involutions are revised. Moreover, in Section~\ref{Sec:QA}, the augmented approach for dealing with quaternions is presented. In Section~\ref{Sec:AQS}, second-order statistics of quaternion-valued random processes, such as correlation and autocorrelation functions, are presented and the reader is introduced to a comprehensive framework for statistical analysis of quaternion-valued random processes. Based on the augmented approach and statistics of quaternion-valued signals, the widely linear model and minimum mean square error (MMSE) estimator are derived in Section~\ref{Sec:MMSE}. In Section~\ref{Sec:CalQ}, the $\mathbb{HR}$-calculus is presented and the reader is introduced to important tools of the $\mathbb{HR}$-calculus. Finally, in Section~\ref{Sec:CalQ}, the reader is introduced to the framework for deriving quaternion-valued adaptive learning techniques.


\section{Quaternion Algebra}

\label{Sec:QA}

On the most fundamental level, a quaternion variable $q \in\mathbb{H}$, consists of a real segment, $\Re(q)$, and a three-dimensional imaginary segment, $\Im(q)$, so that we can write
\begin{equation}
q=\Re\{q\}+\Im\{q\}
\end{equation}
In order to express $\Im\{q\}$ in the imaginary subspace of $\mathbb{H}$, three orthonormal imaginary units that span this space need to be selected. They are generally denoted as $\imath$, $\jmath$, and $\kappa$. These imaginary orthonormal units admit the following product rules~\cite{HamiltonPaper}
\begin{equation}
\imath\jmath=\kappa,\hspace{0.12cm}\jmath\kappa=\imath,\hspace{0.12cm}\kappa\imath=\jmath,\hspace{0.12cm}\imath^{2}=\jmath^{2}=\kappa^{2}=\imath\jmath\kappa=-1
\label{eq:ProductRule}
\end{equation}
In this setting, one can express $q\in\mathbb{H}$ as
\begin{equation}
q =\Re\{q\}+ \Im\{q\}=\Re\{q\} + \Im_{\imath}\{q\} + \Im_{\jmath}\{q\} + \Im_{\kappa}\{q\}=q_{r} + \imath q_{\imath} + \jmath q_{\jmath} + \kappa q_{\kappa}
\label{eq:OriginalQuaternionRep}
\end{equation}
with $\{q_{r},q_{\imath},q_{\jmath},q_{\kappa}\}\in\mathbb{R}$. It is important to note that the selection of imaginary units spanning the imaginary subspace of $\mathbb{H}$ is not unique. Indeed, in some cases, a change in the three imaginary orthonormal units is used to simplify the analysis. We will demonstrate this later in the chapter when presenting derivative rules. 

As a direct result of the relations in \eqref{eq:ProductRule} quaternion product operations become non-commutative, but under exceptional circumstances. To better demonstrate this point, let us multiply both sides of $
\imath\jmath\kappa=-1$
by $\imath$ and carry-out the substitution $\kappa = \imath\jmath$ from \eqref{eq:ProductRule} to arrive at
\begin{equation}
\imath\imath\jmath\imath\jmath=-\imath
\end{equation}
where replacing $\imath^{2}=-1$ from \eqref{eq:ProductRule} yields
\begin{equation}
\jmath\imath\jmath=\imath
\label{eq:ProductProof1}
\end{equation}
Now, multiplying both sides of \eqref{eq:ProductProof1} by $\jmath$ gives
\begin{equation}
\jmath\imath\jmath\jmath=\imath\jmath
\end{equation}
where replacing $\jmath^{2}=-1$ and $\imath\jmath=\kappa$ from \eqref{eq:ProductRule} yields
\begin{equation}
\jmath\imath=-\kappa
\end{equation}
that is a different result from $\imath\jmath=\kappa$ which was given in \eqref{eq:ProductRule}.

More generally, the product of $q_{1}, q_{2}\in\mathbb{H}$ can be found through component-wise multiplication of their constituent elements and using the product rule in \eqref{eq:ProductRule} resulting in
\begin{align}
q_{1}q_{2}=&\Re\{q_{1}\}\Re\{q_{2}\}
\\
&+\Im_{\imath}\{q_{1}\}\Im_{\imath}\{q_{2}\}+\Im_{\jmath}\{q_{1}\}\Im_{\jmath}\{q_{2}\}+\Im_{\kappa}\{q_{1}\}\Im_{\kappa}\{q_{2}\}\nonumber
\\
&+\Re\{q_{1}\}\Im_{\imath}\{q_{2}\}+\Im_{\imath}\{q_{1}\}\Re\{q_{2}\}+\Im_{\jmath}\{q_{1}\}\Im_{\kappa}\{q_{2}\}+\Im_{\kappa}\{q_{1}\}\Im_{\jmath}\{q_{2}\}\nonumber
\\
&+\Re\{q_{1}\}\Im_{\jmath}\{q_{2}\}+\Im_{\imath}\{q_{1}\}\Im_{k}\{q_{2}\}+\Im_{\jmath}\{q_{1}\}\Re\{q_{2}\}+\Im_{\kappa}\{q_{1}\}\Im_{\imath}\{q_{2}\}\nonumber
\\
&+\Re\{q_{1}\}\Im_{k}\{q_{2}\}+\Im_{\imath}\{q_{1}\}\Im_{\jmath}\{q_{2}\}+\Im_{\jmath}\{q_{1}\}\Im_{\imath}\{q_{2}\}+\Im_{\kappa}\{q_{1}\}\Re\{q_{2}\}\nonumber
\end{align}
which can also be formulated using inner product, $\langle\cdot,\cdot\rangle$, and cross product, $\times$, in the imaginary subspace of $\mathbb{H}$ as
\begin{equation}
\begin{aligned}
q_{1}q_{2}=&\Re\{q_{1}\}\Re\{q_{2}\}+\Re\{q_{1}\}\Im\{q_{2}\}+\Re\{q_{2}\}\Im\{q_{1}\} 
\\
&+\Im\{q_{1}\}\times\Im\{q_{2}\}+\big\langle\Im\{q_{1}\},\Im\{q_{2}\}\big\rangle
\end{aligned}
\label{eq:Ch2-Multiplication}
\end{equation}
where
\begin{align}
\big\langle\Im\{q_{1}\},\Im\{q_{2}\}\big\rangle=&\Im_{\imath}\{q_{1}\}\Im_{\imath}\{q_{2}\}+\Im_{\jmath}\{q_{1}\}\Im_{\jmath}\{q_{2}\}+\Im_{\kappa}\{q_{1}\}\Im_{\kappa}\{q_{2}\}
\\
\Im(q_{1})\times\Im(q_{2})=&\text{det}\left(\begin{bmatrix}\imath&\jmath&\kappa\\-\imath\Im_{\imath}\{q_{1}\}&-\jmath\Im_{\jmath}\{q_{1}\}&-\kappa\Im_{\kappa}\{q_{1}\}\\-\imath\Im_{\imath}\{q_{2}\}&-\jmath\Im_{\jmath}\{q_{2}\}&-\kappa\Im_{\kappa}\{q_{2}\}\end{bmatrix}\right)
\end{align}
define the inner and cross product over the three-dimensional imaginary quaternion sub-space. From \eqref{eq:Ch2-Multiplication} note that if $q_{1}$ and $q_{2}$ have parallel imaginary segments, i.e. $\exists\alpha\in\mathbb{R}:\hspace{0.12cm}\Im\{{q}_{1}\}=\alpha\Im\{q_{2}\}$; then, the cross product tends to zeros, that is $\Im(q_{1})\times\Im(q_{2})=0$, and thus, the product becomes commutative. In essence, under the stated condition, $q_{1}$ and $q_{2}$ lie in a subspace of $\mathbb{H}$ spanned by the real axis and $\Im\{q_{1}\}$ (or equivalently $\Im\{q_{2}\}$). Since $\Im^{2}\{q_{1}\}\in(-\infty,0)$, this subspace admits a complex-valued algebra making multiplication commutative. Finally, it is needless to say that for the case where $q_{1}$ and/or $q_{2}$ have vanishing imaginary parts, that is $\Im\{q_{1}\}=0$ and/or $\Im\{q_{2}\}=0$, $\Im(q_{1})\times\Im(q_{2})=0$ making the product commutative. 

The quaternion conjugate is defined as
\begin{equation}
q^{*}\triangleq\Re(q)-\Im(q)
\end{equation}
which allows the second-order norm, hereafter referred to as norm of $q$ for conciseness, to be defined as
\begin{equation}
\left\|q\right\|=\sqrt{qq^{*}}=\sqrt{ q^{2}_{r} + q^{2}_{\imath} + q^{2}_{\jmath} + q^{2}_{\kappa}}
\end{equation}
In turn, this allows an inverse to be defined as
\begin{equation}
\forall q\in\mathbb{H}\backslash\{0\}: q^{-1} = \frac{q^{*}}{\left\|q\right\|^{2}}
\label{eq:MultiplicativeInverse}
\end{equation}
In general, for every two quaternion-valued numbers, $q_{1}$ and $q_{2}$, the following properties follow from the definitions of norm and conjugate
\begin{equation}
\begin{aligned}
\left\|q_{1}q_{2}\right\|=&\left\|q_{1}\right\|\left\|q_{2}\right\|
\\
\left\|\frac{q_{1}}{q_{2}}\right\|=&\frac{\left\|q_{1}\right\|}{\left\|q_{2}\right\|}\hspace{0.12cm}\text{where $q_{2}\neq0$}
\\
(q_{1}q_{2})^{*}=&q^{*}_{2}q^{*}_{1}
\end{aligned}
\label{eq:ConjugateRules}
\end{equation}
As a direct result of non-commutativity of quaternion products, quaternions do not constitute a field.\footnote{On the most elementary level, a field is set on which addition and multiplication operations can be defined in a manner that satisfies the field axioms (associativity, commutativity, distributively, presence of identity, and presence of an inverses) for both addition and multiplication operations. For more details on abstract algebras we refer the keen reader to~\cite{AbstractAlgebra}.} However, given that quaternions admit multiplicative inverse, that is an inverse over $\mathbb{H}\backslash\{0\}$ as defined in \eqref{eq:MultiplicativeInverse}, quaternions admit the definition of a group under multiplication and constitute a division ring (also known as a skew field)~\cite{QuatBook}.\footnote{A division ring, also referred to as a skew field, is a ring in which every non-zero element has a two-sided inverse~\cite{QuatBook}.} For consistency, hereafter, operations on vectors and matrices as pertaining to their scalar elements are carried out in the same order as the vectors and matrices are written. For example, 
\begin{equation}
\text{for}\hspace{3pt}a_{1},a_{2},b_{1},b_{2}\in\mathbb{H}:\hspace{3pt}\begin{bmatrix}a_{1}&a_{2}\end{bmatrix}\begin{bmatrix}b_{1}\\b_{2}\end{bmatrix}=a_{1}b_{1}+a_{2}b_{2}
\end{equation}

In a similar fashion to complex numbers, a quaternion $q \in \mathbb{H}$ can alternatively be expressed by its polar presentation, given by~\cite{QFFT}
\begin{equation}
q = \|q\|e^{\xi \theta}=\|q\|\big(\text{cos}(\theta) + \xi \text{sin}(\theta) \big)
\label{eq:PolarPresentation}
\end{equation}
where
\[
\xi = \frac{\Im(q)}{\|\Im(q)\|}\hspace{1cm}\text{and}\hspace{1cm}\theta = \text{atan}\left(\frac{\|\Im(q)\|}{\Re(q)}\right)
\]
Moreover, it is straightforward to prove that the $\text{sin}(\cdot)$ and $\text{cos}(\cdot)$ functions can be expressed as
\begin{equation}
\text{sin}(\theta)=\frac{1}{2\xi}\left(e^{\xi \theta} - e^{- \xi \theta}\right)\hspace{0.5cm}\text{and}\hspace{0.5cm} \hspace{0.5em}\text{cos}(\theta)=\frac{1}{2}\left(e^{\xi \theta} + e^{- \xi \theta}\right)
\label{eq:QuaternionSinCos}
\end{equation}
where $\xi^{2} = -1$. Note that the real axes and $\xi$ denote a plane in the quaternion domain. As $\xi^{2}=-1$ this plane denotes a complex-valued algebra which yields the formulations of $\sin(\cdot)$ and $\cos(\cdot)$ functions in \eqref{eq:QuaternionSinCos}. Thus, the formulations in \eqref{eq:QuaternionSinCos} are not unique, as $\xi$ can be replaced with an arbitrary quaternion, $\xi'$, as long as $\xi'^{2}=-1$~\cite{PouriaPhD}.



\subsection{Three-Dimensional Rotations}

\begin{figure}[h!]
\centering
\includegraphics[width=1\linewidth,trim=0cm 0cm 0cm 0cm]{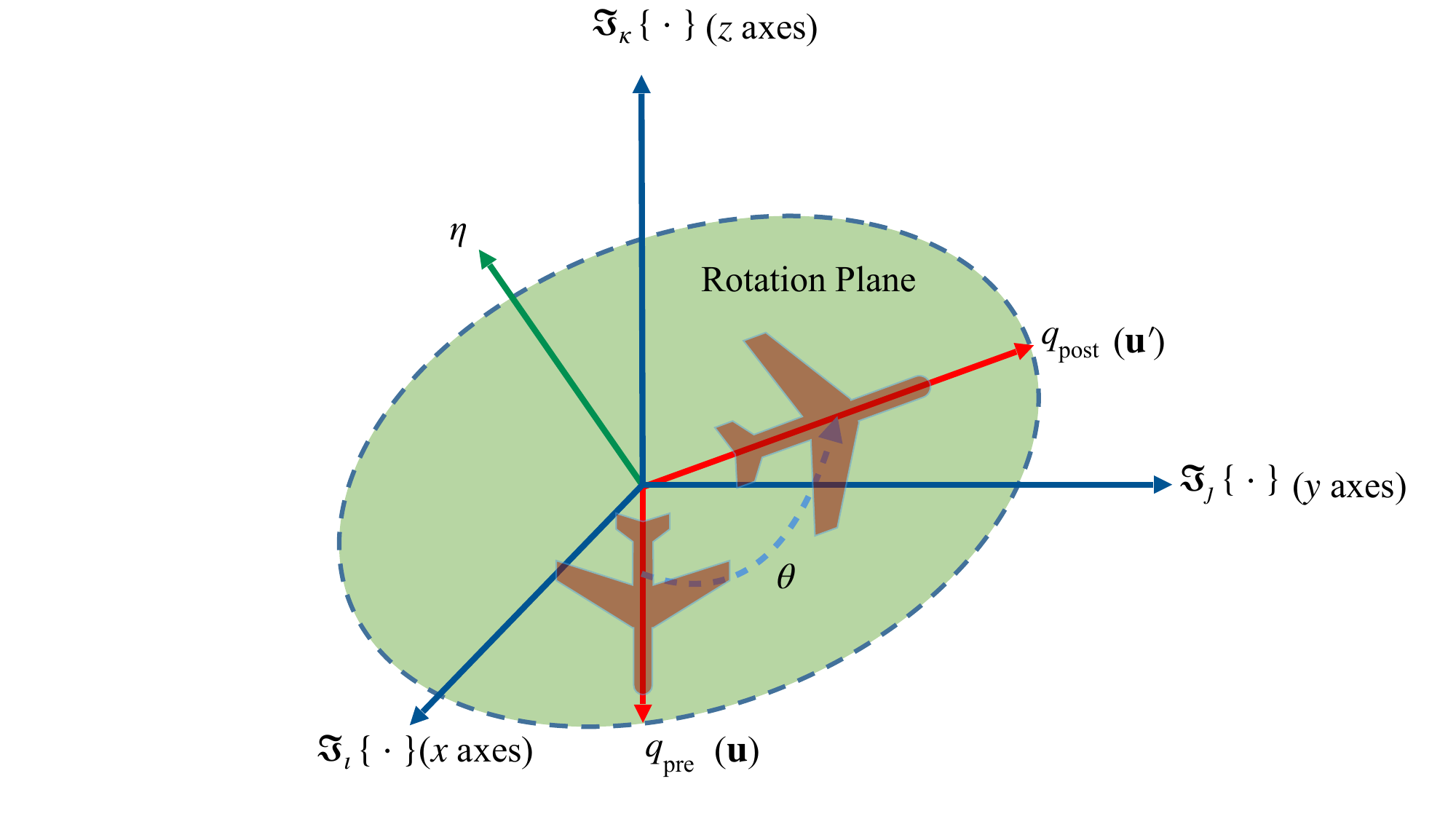}
\caption{Schematic of a rotation around $\eta$ by an angle of $\theta$, with $q_{\text{pre}}$ and $q_{\text{post}}$ pointing to the pre- and post-rotation orientation of the object in question.}
\label{Fig:Rotation}
\end{figure}

The polar presentation of quaternions in \eqref{eq:PolarPresentation} allows for rotations in three-dimensional spaces to be modelled using the quaternion division algebra, which has become perhaps the best known application of quaternions. To this end, consider a rigid body that is put through a rotation. The attitude change of this body can be presented in the Cartesian coordinate system using a single right-hand rotation of that body by an angle of $\theta$ degrees about an axis $\eta$ parallel to the direction that is unchanged by the rotation~\cite{CalssicalRotation}. Moreover, as shown in Figure~\ref{Fig:Rotation}, the three-dimensional Cartesian coordinate system can be considered as representing the imaginary sub-space of a quaternion skew field, so that $\eta=\left(\eta_{x},\eta_{y},\eta_{z}\right)$ can be presented as $\eta=\imath\eta_{x}+\jmath\eta_{y}+\kappa\eta_{z}$. In this setting, the vector presetting the pre- and post-rotation orientation of the object, $q_{\text{pre}}$ and $q_{\text{post}}$, are related as 
\begin{equation}
q_{\text{post}}=\mu q_{\text{pre}}\mu^{-1}\hspace{0.36cm}\text{with}\hspace{0.36cm}\mu=e^{\eta\frac{\theta}{2}}=\cos\left(\frac{\theta}{2}\right)+\eta\sin\left(\frac{\theta}{2}\right)
\label{eq:QuaternionRotation}
\end{equation}

The advantages of modelling three-dimensional rotations employing quaternions as compared to rotation matrices are summarized in the following~\cite{QREV,CG,Kuipers}:

\begin{itemize}

\item A rotation matrix, $\mathbf{R}$, derived based on Euler's theorem must satisfy the unitary condition $\mathbf{R}\mathbf{R}^{\T}=\mathbf{I}$. Therefore, due to finite precision of computer calculations, the need might arise to calibrate the rotation matrix after many rotations have been performed, which is computationally expensive, whereas no such operation is required when modelling rotations with quaternions~\cite{PouriaPhD}.

\item Expressing a sequence of rotations applying the roll, pitch, and yaw angles, a degree of freedom is lost when one of the angles reaches $\pi/2$, referred to as an algebraic gimbal lock. However, this is not the case for quaternions where only the angle and the axis of rotation are required.

\item It is straightforward to produce smooth interpolations of rotations when they are modelled with quaternions allowing for higher quality computer graphics~\cite{CG}.

\end{itemize}

\noindent A detailed trade-off between quaternion model for rotations and other approaches for modelling rotations in three dimensions can be found in~\cite{QREV}.

\subsection{Quaternion Involutions}

\begin{figure}[h]
\centering
\includegraphics[width=1\linewidth,trim = 0cm 0cm 0cm 0cm]{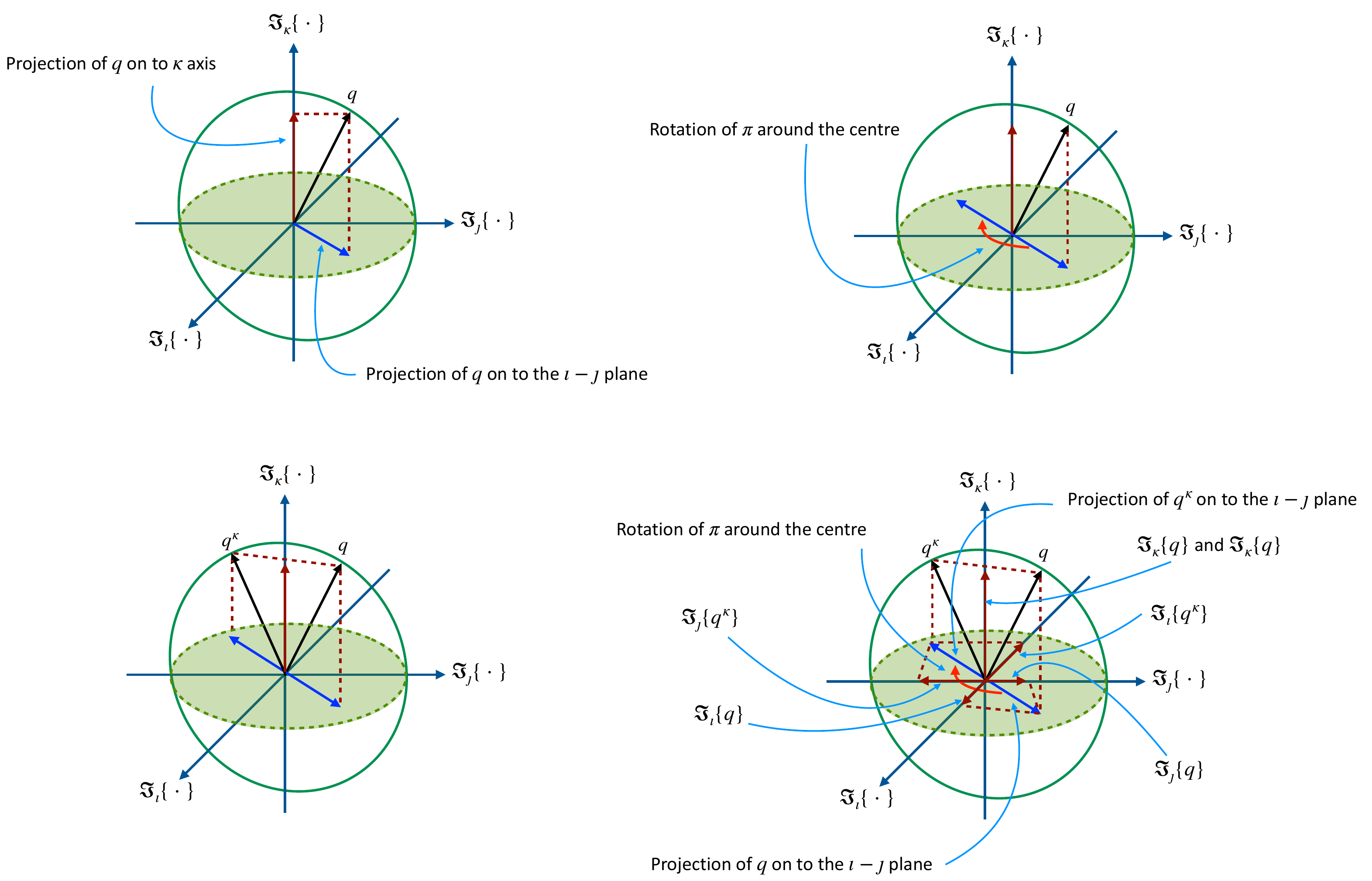}
\caption{Schematic demonstrating $q\in\mathbb{H}$ along side its involution $q^{\kappa}$. The real axis has been omitted for simplicity as $\Re\{q\}$ does not change under quaternion involutions. In the top left graph, $q$ (in black), its projection onto the $\imath-\jmath$ plane (in blue), and its projection onto the $\kappa$ axis (in red) is shown. In the top right graph, projection of $q$ onto the $\imath-\jmath$ plane is rotated around the centre by $\pi$. In the bottom left graph, $q^{\kappa}$ is constructed from the projection of $q$ onto the $\kappa$ axis and the rotated projection of $q$ onto the $\imath-\jmath$ plane. In the bottom right graph, all elements of $q$ and $q^{\kappa}$ are shown together.}
\label{fig:InnvoQI}
\end{figure}

Let us start with recalling the expansion for rotations given in \eqref{eq:QuaternionRotation} for the case of $\theta=\pi$ that results in 
\begin{equation}
\mu=e^{\eta\frac{\pi}{2}}=\cos\left(\frac{\pi}{2}\right)+\eta\sin\left(\frac{\pi}{2}\right)=\eta
\label{eq:RotatePI}
\end{equation}
where $\eta$ is a unit pure imaginary quaternion, that is $\Re\{\eta\}=0$ and $\|\eta^{2}\|=1$ (resulting in $\eta^{2}=-1$). Furthermore, let us implement the operation in \eqref{eq:QuaternionRotation} on a general quaternion number $q\in\mathbb{H}$. In this setting, we have
\begin{equation}
\begin{aligned}
\mu q\mu^{-1} = & \eta\left(\Re\{q\}+\Im\{q\}\right)\eta^{-1}=\eta\Re\{q\}\eta^{-1}+\eta\Im\{q\}\eta^{-1}
\\
= & \Re\{q\}+\eta\Im\{q\}\eta^{-1}
\end{aligned}
\label{eq:InterRotation}
\end{equation}
where we have used \eqref{eq:RotatePI} and $\eta\Re\{q\}\eta^{-1}=\eta\eta^{-1}\Re\{q\}=\Re\{q\}$. In essence, the operation in \eqref{eq:InterRotation} rotates the imaginary segment of $q$ around $\eta$ by $\pi$ while leaving the real segment of $q$ unchanged. This is visualized in Figure~\ref{fig:InnvoQI} for the case of involution around $\kappa$. This is somewhat reminiscent of the complex conjugate operation. Under complex conjugate operation, the real segment of a complex number will remain unchanged, while it can be interpreted that the imaginary segment is rotated around the origin by $\pi$. Indeed, it is this rotation that allows for $z\in\mathbb{C}$ and its conjugate $z^{*}$ to provide two different perspectives of the complex-valued number $z$. Thus, accommodating for i) the real and imaginary components of complex-valued numbers to be separable under simple linear manipulations, ii) the covariance and pseudo-covariance to hold different statistical information, e.g. see \eqref{eq:ComplexCV}-\eqref{eq:ComplexPCV}, iii) treating $z\in\mathbb{C}$ and its conjugate as algebraically independent~\cite{DB,Ken,CI}.\footnote{From algebraic independence, we refer to the fact that for $z\in\mathbb{C}$ we have $\frac{\partial z}{\partial z^{*}}=\frac{\partial z^{*}}{\partial z}=0$~\cite{Ken}.} We seek to provide similar functionality to the complex-conjugate in the quaternion realm. 

Strictly speaking, an involution is defined in mathematics as a mapping that is its own inverse. However, when it comes to quaternions, involutions depict a specific set of mappings~\cite{Q-Invo}. For the purposes of this chapter, we consider quaternion involution, hereafter referred to as involutions for conciseness, to be defined as
\begin{equation}
\forall q,\zeta\in \mathbb{H}\hspace{1pt}\hspace{0.12cm}\text{and}\hspace{0.12cm}\zeta\neq0:q^{\zeta}\triangleq\zeta q \zeta^{-1}
\label{eq:InvoQ}
\end{equation}
which we refer to as involution of $q$ around $\zeta$. Now, consider the following involutions around $\{\imath,\jmath,\kappa\}$ given by~\cite{Q-Invo}
\begin{equation}
\begin{aligned}
q^{\imath} = & q_{r} + \imath q_{\imath} - \jmath q_{\jmath} -\kappa q_{\kappa} 
\\
q^{\jmath} =&q_{r} - \imath q_{\imath} + \jmath q_{\jmath} - \kappa q_{\kappa}
\\
q^{\kappa} =& q_{r} - \imath q_{\imath} - \jmath q_{\jmath} + \kappa q_{\kappa}
\end{aligned}
\label{eq:BasicInvolutions}
\end{equation}
The mappings in \eqref{eq:BasicInvolutions} allow the real-valued components of quaternions to be expressed using the linear operations~\cite{AQS}
\begin{equation}
\begin{aligned}
q_{r} =& \frac{1}{4}\left(q + q^{\imath} + q^{\jmath} + q^{\kappa} \right) & q_{\imath} =&\frac{1}{4\imath}\left(q + q^{\imath} - q^{\jmath} - q^{\kappa}\right)
\\
q_{\jmath} =& \frac{1}{4\jmath}\left(q - q^{\imath}+ q^{\jmath} - q^{\kappa} \right) & q_{\kappa} =&\frac{1}{4\kappa}\left(q - q^{\imath} - q^{\jmath} + q^{\kappa}\right)
\end{aligned}
\label{eq:RealValuedComponents}
\end{equation}
Furthermore, the quaternion conjugate can be formulated as
\begin{equation}
q^{*}=\Re(q)-\Im(q)=\frac{1}{2}\left(q^{\imath} + q^{\jmath} + q^{\kappa} - q \right)
\label{eq:quaternion-conjugate}
\end{equation}

\subsection{The Augmented Quaternion Approach}

In this section, we aim to construct a framework for processing quaternion-valued signals with the following crucial characteristics: 

\begin{itemize}
	
\item The framework must be constructed so that most concepts in real-valued statistical analysis and calculus, such as covariance, expectation, derivatives, and direction of steepest descent,  can be extended to the quaternion domain in an intuitive and straightforward manner.
	
\item The framework must be able to preserve all calculations and mathematical operations within the quaternion division algebra.
	
\end{itemize}

\noindent In order to achieve the first goal, we need a basis to relate a quaternion-valued signal to its real-valued components. To this end, we start with defining the involution operation on a quaternion vector as the element-wise implementation of that involution on the components of the quaternion vector, that is, for $\{q_{1},\ldots,q_{n}\}\in\mathbb{H}$ and $\zeta\in\mathbb{H}$, we have
\begin{equation}
\mathbf{q}^{\zeta}= \left(\begin{bmatrix}q_{1}\\\vdots\\q_{n}\end{bmatrix}\right)^{\zeta}=\left(\begin{bmatrix}q^{\zeta}_{1}\\\vdots\\q^{\zeta}_{n}\end{bmatrix}\right)
\end{equation}
Now, using the relation in \eqref{eq:RealValuedComponents} we can formulate the relation between $\mathbf{q}$ and its real-valued components as~\cite{AQS}
\begin{equation}
\begin{bmatrix}\mathbf{q}_{r}\\\mathbf{q}_{\imath}\\\mathbf{q}_{\jmath}\\\mathbf{q}_{\kappa}\end{bmatrix}=\frac{1}{4}\begin{bmatrix}\phantom{-l}\mathbf{I}&\phantom{-l}\mathbf{I}&\phantom{-l}\mathbf{I}&\phantom{-l}\mathbf{I}\\-\imath\mathbf{I}&-\imath\mathbf{I}&\phantom{-}\imath\mathbf{I}&\phantom{-}\imath\mathbf{I}\\-\jmath\mathbf{I}&\phantom{-}\jmath\mathbf{I}&-\jmath\mathbf{I}&\phantom{-}\jmath\mathbf{I}\\-\kappa\mathbf{I}&\phantom{-}\kappa\mathbf{I}&\phantom{-}\kappa\mathbf{I}&-\kappa\mathbf{I}\end{bmatrix}\begin{bmatrix}\mathbf{q}\\\mathbf{q}^{\imath}\\\mathbf{q}^{\jmath}\\\mathbf{q}^{\kappa}\end{bmatrix}
\label{eq:MappingHRinRev}
\end{equation}
and hence
\begin{equation}
\begin{bmatrix}\mathbf{q}^{\phantom{l}}\\\mathbf{q}^{\imath}\\\mathbf{q}^{\jmath}\\\mathbf{q}^{\kappa}\end{bmatrix}=\begin{bmatrix}\mathbf{I}&\phantom{-}\imath\mathbf{I}&\phantom{-}\jmath\mathbf{I}&\phantom{-}\kappa\mathbf{I}\\\mathbf{I}&\phantom{-}\imath\mathbf{I}&-\jmath\mathbf{I}&-\kappa\mathbf{I}\\\mathbf{I}&-\imath\mathbf{I}&\phantom{-}\jmath\mathbf{I}&-\kappa\mathbf{I}\\\mathbf{I}&-\imath\mathbf{I}&-\jmath\mathbf{I}&\phantom{-}\kappa\mathbf{I}\end{bmatrix}\begin{bmatrix}\mathbf{q}_{r}\\\mathbf{q}_{\imath}\\\mathbf{q}_{\jmath}\\\mathbf{q}_{\kappa}\end{bmatrix}
\label{eq:MappingHR}
\end{equation}
where $\mathbf{q}_{r}=\begin{bmatrix}q_{1_{r}},\ldots,q_{n_{r}}\end{bmatrix}^{\T}$ with $\mathbf{q}_{\imath}$, $\mathbf{q}_{\jmath}$, $\mathbf{q}_{\kappa}$ defined similarly. Note that \eqref{eq:MappingHRinRev} follows directly from generalising the relations in \eqref{eq:RealValuedComponents} into a vector format.

Perhaps, the most important step in the formulation of the augmented approach is using $\{\mathbf{q}^{\imath},\mathbf{q}^{\jmath},\mathbf{q}^{\kappa}\}$ alongside $\mathbf{q}$ in order to establish a linear operator that maps $\mathbf{q}$ to its real-valued components. In essence, the involutions of $\mathbf{q}$ around the orthonormal axes $\imath$, $\jmath$, and $\kappa$, allow the four-dimensional quaternion $\mathbf{q}$ to offer four different perspectives that pose complementary information. This is a crucial concept when it comes to dealing with quaternions. We therefore denote the augmented vector as $\mathbf{q}^{a}$ and its mapping $\mathbf{A}$ to its real-valued counterpart as
\begin{equation}
\mathbf{q}^{a}=\begin{bmatrix}\mathbf{q}^{\phantom{l}}\\\mathbf{q}^{\imath}\\\mathbf{q}^{\jmath}\\\mathbf{q}^{\kappa}\end{bmatrix}\hspace{0.32cm}\text{and}\hspace{0.32cm}\mathbf{A}=\begin{bmatrix}\mathbf{I}&\phantom{-}\imath\mathbf{I}&\phantom{-}\jmath\mathbf{I}&\phantom{-}\kappa\mathbf{I}\\\mathbf{I}&\phantom{-}\imath\mathbf{I}&-\jmath\mathbf{I}&-\kappa\mathbf{I}\\\mathbf{I}&-\imath\mathbf{I}&\phantom{-}\jmath\mathbf{I}&-\kappa\mathbf{I}\\\mathbf{I}&-\imath\mathbf{I}&-\jmath\mathbf{I}&\phantom{-}\kappa\mathbf{I}\end{bmatrix}
\label{eq: augmented basis}
\end{equation}
where the mapping $\mathbf{A}$ can be straightforwardly inverted as
\begin{equation}
\mathbf{A}^{-1}=\frac{1}{4}\mathbf{A}^{\H}
\end{equation}
This augmented basis is used in Section \ref{Sec:MMSE} to derive a framework for quaternion calculus. Prior to the introduction of the quaternion calculus, next section shows how the augmented basis enables the capture of complete statistical information of the quaternion space in the context of \emph{sufficient} statistics.


\section{Augmented Statistics}
\label{Sec:AQS}
\begin{figure}[ht]
     \adjustimage{width=1.4\textwidth,center}{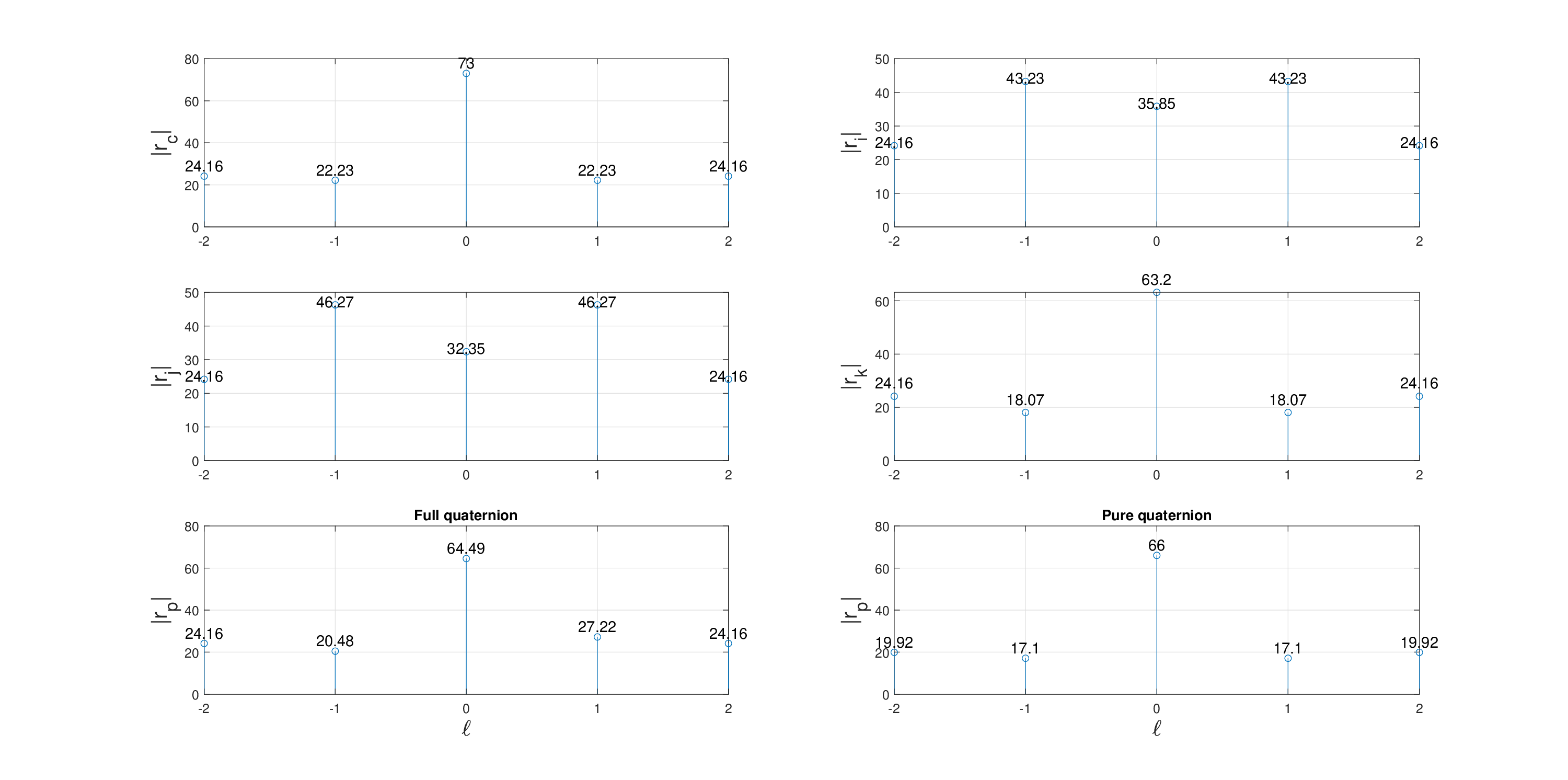}
    \caption{Absolute magnitude of autocorrelations as a function of lag. Bottom row: The left plot shows the non-symmetric pseudo-autocorrelation of full quaternion sequence, whereas the right plot shows its symmetric pseudo-autocorrelation, if its imaginary part was considered as pure quaternions.}
    \label{fig: autocorrelation01}
\end{figure}

As the autocorrelation $r_c(\ell)$ is widely used in signal processing, it is considered in this chapter to illustrate how the quaternion involutions in (\ref{eq:BasicInvolutions}) can \emph{augment} the second order statistics beyond the traditional definition of the autocorrelation, which is given by
\begin{align}
r_c(\ell)&=\E{q(n)q^*(n-\ell)}
\label{eq: autocorrelation c}
\end{align}
Recall that the autocorrelation $r_c(\ell)$ is conjugate symmetric, i.e. $r_c(\ell)=r_c^*(-\ell)$. Moreover, the autocorrelation $r_c(\ell)$ reaches its maximum at $\ell=0$, i.e. $r_c(0)\geq r_c(\ell)~\forall\ell\neq 0$. The quaternion involutions in (\ref{eq:BasicInvolutions}) mean that it is no longer sufficient to consider only the autocorrelation in (\ref{eq: autocorrelation c}). In other words, the second order statistics of quaternion variables need to be augmented with the $\eta$-autocorrelations $r_\eta(\ell)$, which can be formulated as 
\begin{align}
       r_\eta(\ell)&=\E{q(n)q^{\eta *}(n-\ell)} \label{eq: autocorrelation ijk}
\end{align}
Note that the $\eta$-autocorrelation $r_\eta(\ell)$ is conjugate $\eta$-symmetric, i.e. $r_\eta(\ell)=r_\eta^{\eta *}(-\ell)$.  For completeness, the pseudo-autocorrelation $r_p(\ell)$ is also defined in $\mathbb{H}$ as
\begin{align}
       r_p(\ell)&=\E{ q(n)q(n-\ell)} \label{eq: autocorrelation p}
\end{align}
Finally,  all autocorrelations $r(\ell)$ in (\ref{eq: autocorrelation c})-(\ref{eq: autocorrelation p}) are symmetric in terms of the absolute magnitude, i.e. $|r(\ell)|=|r(-\ell)|$ for \emph{pure}\footnote{For full quaternions, their absolute magnitudes are also symmetric except for $r_p(\ell)$.} quaternions. See Figure \ref{fig: autocorrelation01} for illustration. Moreover, these autocorrelations are related as follows
       \begin{align}
     r_p(\ell)&=\frac{1}{2}\bigg(r_\imath(\ell)+r_\jmath(\ell)+r_\kappa(\ell)-r_c(\ell) \bigg)
     \label{eq: correlation dependency}
     \end{align}
 As one of these autocorrelations can be inferred from the other four autocorrelations, any four of these autocorrelations provide sufficient second order statistics, in general. There are special cases, however, when only two or three of these autocorrelations offer sufficient statistics, see \cite{CliveSPM2024}.
%
\begin{example}\label{Ex: Autocorrelation 01}
%
    \begin{equation}
\mathbf{q}(n) =\bigg[-1-10\imath+\jmath-\kappa,~{-2-4\imath-6\jmath+3\kappa},~-4-5\imath+3\jmath+\kappa \bigg]\label{eq: example quaternion sequence}
\end{equation}
\end{example}
\\The following MATLAB code shows the computation of the sample estimate of $\imath$-autocorrelation. The autocorrelations of (\ref{eq: example quaternion sequence}) are provided in (\ref{eq: rc})-(\ref{eq: rpFull}).
%
\begin{lstlisting}[
frame=single,
numbers=left,
style=Matlab-editor, basicstyle=\footnotesize]
%Quaternion examplar sequence
q=[quaternion(-1,-10,1,-1) quaternion(-2,-4,-6,3) quaternion(-4,-5,3,1)];
%Computation of i-autocorrelation
r_i=conv(q,conj(fliplr(invijk(q,'i'))))/length(q);
%Plot the absolute i-autocorrelation for visual symmetricity
figure,stem(k,abs(r_i)),grid
ylabel('|r_i|')
xlabel('$\mathbf{\ell}$','interpreter','latex')
\end{lstlisting}
Example \ref{Ex: Autocorrelation 01} uses the quaternion toolbox \cite{sangwine05} and the function invijk(q,`i') in line 4 implements the $\imath-$involution, which can be downloaded at \href{https://www.researchgate.net/publication/327765279_Quaternion_Statistics}{here}.
\begin{align}
\mathbf{r}_c
&=\bigg[r_c(-2),r_c(-1),~\underset{\mathlarger{\mathbf{\uparrow}}}{r_c(0)} , r_c(1),r_c(2) \bigg]\nonumber\\
&=\bigg[18.67+10.33\imath-5.33\jmath+10\kappa,~15.33+13.33\imath-0.33\jmath-9\kappa,~\underset{\mathlarger{\mathbf{\uparrow}}}{73},\nonumber\\
&~15.33-13.33\imath+0.33\jmath+9\kappa,~18.67-10.33\imath+5.33\jmath-10\kappa \bigg]\label{eq: rc}
\end{align}
\begin{align}
%
\mathbf{r}_\imath
&=\bigg[r_\imath(-2),r_\imath(-1),~\underset{\mathlarger{\mathbf{\uparrow}}}{r_\imath(0)}, r_\imath(1),r_\imath(2) \bigg]\nonumber\\
&=\bigg[17.33+13\imath-0.67\jmath-10.67\kappa,~31.33+1.33\imath+22.33\jmath+19.67\kappa,\nonumber\\
&~\underset{\mathlarger{\mathbf{\uparrow}}}{35+4\jmath-0.67\kappa},\label{eq: ri}\\
&~31.33-1.33\imath+22.33\jmath+19.67\kappa,~17.33-13\imath-0.67\jmath-10.67\kappa \bigg]\nonumber
\end{align}
\begin{align}
%
\mathbf{r}_\jmath
&=\bigg[r_\jmath(-2),r_\jmath(-1),~\underset{\mathlarger{\mathbf{\uparrow}}}{r_\jmath(0)}, r_\jmath(1),r_\jmath(2) \bigg]\nonumber\\
&=\bigg[-14+14.33\imath+4.67\jmath+12.67\kappa,~-24.67+20.67\imath+15\jmath-29.67\kappa,\nonumber\\
&~\underset{\mathlarger{\mathbf{\uparrow}}}{-28.33+14.67\imath-5.33\kappa},\label{eq: rj}\\
&~-24.67+20.67\imath-15\jmath-29.67\kappa,~-14+14.33\imath-4.67\jmath+12.67\kappa \bigg]\nonumber
\end{align}
\begin{align}
%
%
\mathbf{r}_\kappa
&=\bigg[r_\kappa(-2),r_\kappa(-1),~\underset{\mathlarger{\mathbf{\uparrow}}}{r_\kappa(0)}, r_\kappa(1),r_\kappa(2) \bigg]\nonumber\\
&=\bigg[-16.67 + 15.67 \imath-4\jmath-6.67\kappa,~-8.67 + 12.67\imath -7.67 \jmath + 5.67\kappa,
\nonumber\\
&~\underset{\mathlarger{\mathbf{\uparrow}}}{-51.67 + 36 \imath -5.33 \jmath},\label{eq: rk}\\
&~-8.67 + 12.67 \imath -7.67 \jmath -5.67 \kappa,~ -16.67 + 15.67 \imath  -4 \jmath + 6.67 \kappa \bigg]\nonumber
\end{align}
\begin{align}
%
\mathbf{r}_p^{\textrm{Full}} 
&=\bigg[r_p(-2),r_p(-1),~\underset{\mathlarger{\mathbf{\uparrow}}}{r_p(0)}, r_p(1),r_p(2) \bigg]\nonumber\\
&=\bigg[-16 + 16.33\imath + 2.67\jmath-7.33\kappa,~-8.67 + 10.67 \imath + 15\jmath + 2.33 \kappa,
\nonumber\\
&~\underset{\mathlarger{\mathbf{\uparrow}}}{-59 + 25.33\imath -0.67 \jmath -6 \kappa},\label{eq: rpFull}\\
&~ -8.67 + 22.67\imath-0.33\jmath-12.33\kappa,~ -16 + 13.67\imath-7.33 \jmath + 9.33 \kappa \bigg]\nonumber
\end{align}
\begin{align}
%
\mathbf{r}_p^{\textrm{Pure}}
&=\bigg[r_p(-2),r_p(-1),~\underset{\mathlarger{\mathbf{\uparrow}}}{r_p(0)}, r_p(1),r_p(2) \bigg]\nonumber\\
&=\bigg[-17.33 + 1.33 \imath + 5\jmath -8.33\kappa,~-12-6\imath + 7.67\jmath + 7.33\kappa,~\underset{\mathlarger{\mathbf{\uparrow}}}{-66},\nonumber\\
&~ -12 + 6\imath-7.67\jmath-7.33\kappa,~-17.33-1.33\imath-5\jmath+8.33\kappa \bigg]\label{eq: rpPure}
\end{align}
The last equation (\ref{eq: rpPure}) shows the conjugate symmetric pseudo-autocorrelation of (\ref{eq: example quaternion sequence}) if its imaginary part is modelled as a pure quaternion sequence. Figure \ref{fig: autocorrelation01} shows the absolute values of the autocorrelations in (\ref{eq: rc})-(\ref{eq: rpPure}). Observe the symmetries for all autocorrelation sequences, except for $r_p^{\textrm{Full}}(\ell)$.
\\\\Note the conjugate $\eta$-symmetricity in (\ref{eq: ri})-(\ref{eq: rk}). For example, in (\ref{eq: ri}), $r_\imath(-1)= r_\imath^{\imath *}(1)=31.33+1.33\imath+22.33\jmath+19.67\kappa$. In other words, only the $\imath$ imaginary components are conjugate of each other at $\ell=1,-1$.  Moreover, $\imath$-imaginary component vanishes at $\ell=0$. 
\\\\As expected, the pseudo autocorrelation in (\ref{eq: rpFull}) is not symmetric, since $r_p(\ell)\neq r_p(-\ell)$ due to the non-commutativity of the quaternion product. However, if the imaginary parts of quaternion sequence in (\ref{eq: example quaternion sequence}) are considered as pure quaternions, then the pseudo autocorrelation becomes symmetric as in (\ref{eq: rpPure}). In this particular scenario (\ref{eq: rpPure}), $r_p(\ell)=-r_c(\ell)$, see \cite{CliveSPM2024} for further details.
\subsection{The Autocorrelation Matrices}
Following the definition of these autocorrelations, their matrix counterparts can now be introduced. Consider the vector
\begin{equation}
\mathbf{q}(n)= \bigg[q(n),q(n-1),\ldots,q(n-L)\bigg]^\T
\end{equation}
of $L+1$ quaternion values. Then, its outer product is 
\begin{equation} \label{eq: correlation ijk matrix part 1}
           \mathbf{q}(n)\mathbf{q}^{\eta \H}(n)=
        \begin{bmatrix}
            q(n)q^{\eta *}(n) & q(n)q^{\eta *}(n-1) & \cdots & q(n)q^{\eta *}(n-L)\\
            q(n-1)q^{\eta *}(n) & q(n-1)q^{\eta *}(n-1) & \cdots& q(n-1)q^{\eta *}(n-L)\\
            \vdots&\vdots&\ddots&\vdots\\
            q(n-L)q^{\eta *}(n) & q(n-L)q^{\eta *}(n-1) & \cdots & q(n-L)q^{\eta *}(n-L)
        \end{bmatrix}
\end{equation}
which is a $(L+1) \times (L+1)$ matrix. The expected value of (\ref{eq: correlation ijk matrix part 1}) gives the $\eta$-autocorrelation matrix
\begin{equation} \label{eq: correlation ijk matrix part 2}
           \mathbf{R}_{\eta}(n)=
        \begin{bmatrix}
            r_\eta(0) & r_\eta(1) & \cdots & r_\eta(L)\\
            r_\eta(-1) & r_\eta(0) & \cdots& r_\eta(L-1)\\
            \vdots&\vdots&\ddots&\vdots\\
            r_\eta(-L) & r_\eta(1-L) & \cdots & r_\eta(0)
        \end{bmatrix}
        =
            \begin{bmatrix}
            r_\eta(0) & r_\eta(1) & \cdots & r_\eta(L)\\
            r_\eta^{\eta *}(1) & r_\eta(0) & \cdots& r_\eta(L-1)\\
            \vdots&\vdots&\ddots&\vdots\\
            r_\eta^{\eta *}(L) & r_\eta^{\eta *}(L-1) & \cdots & r_\eta(0)
        \end{bmatrix}
\end{equation}
The $\eta$-autocorrelation matrix has a Toeplitz structure, as its elements are equal on each of its diagonals. For example, $r_\eta(0)$ lies on its main diagonal, while the diagonal over its main diagonal have elements equal to $r_\eta(1)$. Another property of $\eta$-autocorrelation matrix is its $\eta$-Hermitian structure, i.e. $\mathbf{R}_{\eta}(n)=\mathbf{R}_{\eta}^{\eta \H}(n)$. Due to this structure, the $\eta$-autocorrelation can be decomposed as \cite{CHEONGTOOKijkHermitian}
    \begin{equation}
        \mathbf{R}_{\eta}(n)=\boldsymbol{\diameter}\boldsymbol{\Lambda}\boldsymbol{\diameter}^{\eta \H}
        \label{eq: factorisation takagi}
    \end{equation}
where $\boldsymbol{\Lambda}$ has the singular values of $\mathbf{R}_{\eta}[n]$ on its main diagonal, and $\boldsymbol{\diameter}$ is a unitary matrix, which can be computed as 
\begin{equation}
\mathbf{\diameter}=\mathbf{U}(\mathbf{D}^\eta)^{1/2},\textrm{ where }\mathbf{D}=\mathbf{V}^{\eta \H}\mathbf{U}
\end{equation}
$\mathbf{U}$ and $\mathbf{V}$ can be obtained from the singular value decomposition of $\mathbf{R}_{\eta}=\mathbf{U}\Lambda\mathbf{V}^\H$. The MATLAB code for the factorisation in (\ref{eq: factorisation takagi}) can be downloaded \href{https://www.researchgate.net/publication/275037850_MATLAB_for_Quaternion_Takagi_factorisation}{here}.
Similarly, the Hermitian autocorrelation $\mathbf{R}(n)$ and the pseudo autocorrelation $\mathbf{R}_{p}(n)$ matrices can be expressed as
\begin{equation} \label{eq: correlation matrix part}
          \mathbf{R}_c(n)=\E{ \mathbf{q}(n)\mathbf{q}^{\H}(n)}=
           \begin{bmatrix}
            r(0) & r(1) & \cdots & r(L)\\
            r^{*}(1) & r_\eta(0) & \cdots& r(L-1)\\
            \vdots&\vdots&\ddots&\vdots\\
            r^{*}(L) & r^{*}(L-1) & \cdots & r(0)
        \end{bmatrix}
\end{equation}
\begin{equation} \label{eq: pcorrelation matrix part}
          \mathbf{R}_{p}(n)=\E{ \mathbf{q}(n)\mathbf{q}^{\T}(n)}=
        \begin{bmatrix}
            r_p(0) & r_p(1) & \cdots & r_p(L)\\
            r_p(-1) & r_p(0) & \cdots& r_p(L-1)\\
            \vdots&\vdots&\ddots&\vdots\\
            r_p(-L) & r_p(1-L) & \cdots & r_p(0)
        \end{bmatrix}
\end{equation}
The non-symmetricity of the pseudo autocorrelation matrix $\mathbf{R}_{p}(n)\neq \mathbf{R}_{p}^\T(n)$ arises as a consequence of $r_p(\ell)\neq r_p(-\ell)$. This also implies that the pseudo autocorrelation matrix \emph{cannot} be factorised as the Takagi factorisation $\mathbf{R}_{p}(n)\neq \mathbf{\Psi}\Lambda\mathbf{\Psi}^\T$ like in the complex domain $\mathbb{C}$ .
\begin{table}
\caption{Quaternion-valued autocorrelation matrices considered in Example \ref{ex: example autocorrelation quaternion matrices}. Observe the Hermitian structure of ${R}_c(n)$ as  shown in (\ref{eq: correlation matrix part}) and the $\eta-$Hermitian property of $\mathbf{R}_{\imath}(n)$, $\mathbf{R}_{\jmath}(n)$, and $\mathbf{R}_{\kappa}(n)$ as illustrated in (\ref{eq: correlation ijk matrix part 2}).}
\label{tab: autocorrelation quaternion matrices}
\begin{tabular}{|*1{>{\centering\arraybackslash}p{1.2\textwidth}|}}
\hline
{Autocorrelation matrices in $\mathbb{H}$ (of size $3 \times 3$)}  \\
\hline
 \[ \mathbf{R}_c(n)=\left[ \begin{array}{ccc}   
        {\color{darkgray} 73}  & 15.33-13.33\imath   & {\color{gray}18.67-10.33\imath}\\ 
                               & +0.33\jmath+9\kappa & {\color{gray}+5.33\jmath-10\kappa}\\
                            15.33+13.33\imath &  {\color{gray} 73} & 15.33-13.33\imath \\
                           -0.33\jmath-9\kappa&                       & +0.33\jmath+9\kappa\\
                        {\color{gray}18.67+10.33\imath}   & 15.33+13.33\imath  &  {\color{gray} 73}\\
                        {\color{gray}-5.33\jmath+10\kappa} & -0.33\jmath-9\kappa   & \end{array}\right]\] 
\\\hline
 \[\mathbf{R}_{\imath}(n)=\left[ \begin{array}{ccc} 
        {\color{gray} 35+4\jmath-6.67\kappa} & 31.33-1.33\imath         & {\color{gray}17.33-13\imath}\\ 
                                                 & +22.33\jmath+19.67\kappa & {\color{gray}-0.67\jmath-10.67\kappa}\\
                            31.33+1.33\imath       &  {\color{gray} 35+4\jmath-6.67\kappa} & 31.33-1.33\imath \\
                           +22.33\jmath-19.67\kappa&                                           & +22.33\jmath+19.67\kappa\\
                 {\color{gray}17.33+13\imath}          & 31.33+1.33\imath          &  {\color{gray} 35+4\jmath-6.67\kappa}\\
                 {\color{gray}-0.67\jmath-10.67\kappa} & +22.33\jmath+19.67\kappa   & \end{array}\right]\] 
\\\hline
 \[\mathbf{R}_{\jmath}(n)=\left[ \begin{array}{ccc} 
        {\color{gray}-28.33+14.67\imath-5.33\kappa} & -24.67+20.67\imath    & {\color{gray}-14+14.33\imath}\\ 
                                                        & -15\jmath-29.67\kappa & {\color{gray}-4.67\jmath+12.67\kappa}\\
                            -24.67+20.67\imath  &  {\color{gray}-28.33+14.67\imath-5.33\kappa} & -24.67+20.67\imath \\
                           +15\jmath-29.67\kappa&                                           & -15\jmath-29.67\kappa\\
        {\color{gray}-14+14.33\imath}         & -24.67+20.67\imath          &  {\color{gray}-28.33+14.67\imath-5.33\kappa}\\
        {\color{gray}+4.67\jmath+12.67\kappa} & +15\jmath-29.67\kappa   & \end{array}\right]\] 
\\\hline
 \[\mathbf{R}_{\kappa}(n)=\left[ \begin{array}{ccc} 
        {\color{gray}-51.67+36\imath-5.33\jmath} & -8.67+12.67\imath    & {\color{gray}-16.67+15.67\imath}\\ 
                                                        & -7.67\jmath-5.67\kappa & {\color{gray}-4\jmath+6.67\kappa}\\
                            -8.67+12.67\imath    & {\color{gray}-51.67+36\imath-5.33\jmath} & -8.67+12.67\imath \\
                           -7.67\jmath+5.67\kappa&                                           & -7.67\jmath-5.67\kappa\\
        {\color{gray}-16.67+15.67\imath}  & -8.67+12.67\imath          &  {\color{gray}-51.67+36\imath-5.33\jmath}\\
        {\color{gray}-4\jmath-6.67\kappa} & -7.67\jmath+5.67\kappa   & \end{array}\right]\] 
\\\hline
\end{tabular}
\end{table}
\begin{example}\label{ex: example autocorrelation quaternion matrices}
    Consider the quaternion sequence in (\ref{eq: example quaternion sequence}). Its augmented second order \emph{multi-dimensional} statistics are summarised in Table \ref{tab: autocorrelation quaternion matrices}. These autocorrelation matrices can be constructed from their corresponding autocorrelation sequences in (\ref{eq: rc})-(\ref{eq: rk}) based on the structures in (\ref{eq: correlation ijk matrix part 2}) and (\ref{eq: correlation matrix part}).
\end{example}
\subsection{Duality between auto-correlation matrices in $\mathbb{R}$ and  $\mathbb{H}$ } 
\noindent The pseudo-autocorrelation matrix $\mathbf{R}_p(n)$ in (\ref{eq: pcorrelation matrix part}) can be expressed as its real-valued autocorrelation matrices as follows
\begin{align}
\mathbf{R}_p(n)&=\Re\{\mathbf{R}_p(n)\}+\imath
\Im_\imath\{\mathbf{R}_p(n)\}+\jmath
\Im_\jmath\{\mathbf{R}_p(n)\}+\kappa
\Im_\kappa\{\mathbf{R}_p(n)\}
\end{align}
\begin{align}
\textrm{where~~~~}
\Re\{\mathbf{R}_p(n)\}&=\mathbf{R}_{rr}(n)-\mathbf{R}_{\imath\imath}(n)-\mathbf{R}_{\jmath\jmath}(n)-\mathbf{R}_{\kappa\kappa}(n)\nonumber\\
\Im_\imath\{\mathbf{R}_p(n)\}&=\mathbf{R}_{r \imath}(n)+\mathbf{R}_{\imath r}(n)+\mathbf{R}_{\jmath \kappa}(n)-\mathbf{R}_{\kappa \jmath}(n)\nonumber\\
\Im_\jmath\{\mathbf{R}_p(n)\}&=\mathbf{R}_{r \jmath}(n)+\mathbf{R}_{\jmath r}(n)-\mathbf{R}_{\imath \kappa}(n)+\mathbf{R}_{\kappa \imath}(n)\nonumber\\
\Im_\kappa\{\mathbf{R}_p(n)\}&=\mathbf{R}_{r
\kappa}(n)+\mathbf{R}_{\kappa
r}(n)-\mathbf{R}_{\jmath
\imath}(n)+\mathbf{R}_{\imath \jmath}(n)\nonumber
~~~~~~~~~~~~~~~~~~~~~~~~~~~~~~~~~~~~~~
\end{align}
%
%
\begin{table}[t]
\caption{{\small Quaternion-valued auto-correlation
matrices in terms of their real-valued
counterparts}} \label{tab: structure covariance1}
\begin{center}
\begin{tabular}{|c|c|c|}\hline
Auto-correlation &
$\mathbf{R}_c(n)$
&$\mathbf{R}_{\imath}(n)$
\\
matrix &
&
\\\hline
$\Re\{\cdot\}$ &
$\mathbf{R}_{rr}(n)+\mathbf{R}_{\imath\imath}(n)+\mathbf{R}_{\jmath\jmath}(n)+\mathbf{R}_{\kappa\kappa}(n)$
& $\mathbf{R}_{rr}(n)+\mathbf{R}_{\imath\imath}(n)-\mathbf{R}_{\jmath\jmath}(n)-\mathbf{R}_{\kappa\kappa}(n)$ \\
$\Im_\imath\{\cdot\}$ & $\mathbf{R}_{\imath
r}(n)-\mathbf{R}_{r
\imath}(n)+\mathbf{R}_{\kappa
\jmath}(n)-\mathbf{R}_{\jmath \kappa}(n)$
& $\mathbf{R}_{\imath r}(n)-\mathbf{R}_{r \imath}(n)+\mathbf{R}_{\jmath \kappa}(n)-\mathbf{R}_{\kappa \jmath}(n)$ \\
$\Im_\jmath\{\cdot\}$ & $\mathbf{R}_{\jmath
r}(n)-\mathbf{R}_{r
\jmath}(n)+\mathbf{R}_{\imath
\kappa}(n)-\mathbf{R}_{\kappa \imath}(n)$ &
$\mathbf{R}_{r \jmath}(n)+\mathbf{R}_{\jmath r}(n)-\mathbf{R}_{\kappa \imath}(n)-\mathbf{R}_{\imath \kappa}(n)$ \\
$\Im_\kappa\{\cdot\}$ & $\mathbf{R}_{\kappa
r}(n)-\mathbf{R}_{r
\kappa}(n)+\mathbf{R}_{\jmath
\imath}(n)-\mathbf{R}_{\imath \jmath}(n)$ &
$\mathbf{R}_{\kappa r}(n)+\mathbf{R}_{r
\kappa}(n)+\mathbf{R}_{\imath
\jmath}(n)+\mathbf{R}_{\jmath \imath}(n)$\\
\hline
\end{tabular}
\end{center}
\end{table}
\begin{table}[t]
\caption{{\small Quaternion-valued auto-correlation
matrices in terms of their real-valued
counterparts}} \label{tab: structure covariance2}
\begin{center}
\begin{tabular}{|c|c|c|}\hline
Auto-correlation &
$\mathbf{R}_{\jmath}(n)$ &
$\mathbf{R}_{\kappa}(n)$
\\
matrix &
&
\\\hline
$\Re\{\cdot\}$ &
$\mathbf{R}_{rr}(n)-\mathbf{R}_{\imath\imath}(n)+\mathbf{R}_{\jmath\jmath}(n)-\mathbf{R}_{\kappa\kappa}(n)$
& $\mathbf{R}_{rr}(n)-\mathbf{R}_{\imath\imath}(n)-\mathbf{R}_{\jmath\jmath}(n)+\mathbf{R}_{\kappa\kappa}(n)$ \\
$\Im_\imath\{\cdot\}$ & $\mathbf{R}_{\imath
r}(n)+\mathbf{R}_{r
\imath}(n)+\mathbf{R}_{\kappa
\jmath}(n)+\mathbf{R}_{\jmath \kappa}(n)$
& $\mathbf{R}_{\imath r}(n)+\mathbf{R}_{r \imath}(n)-\mathbf{R}_{\jmath \kappa}(n)-\mathbf{R}_{\kappa \jmath}(n)$ \\
$\Im_\jmath\{\cdot\}$ & $\mathbf{R}_{\jmath
r}(n)-\mathbf{R}_{r
\jmath}(n)+\mathbf{R}_{\kappa
\imath}(n)-\mathbf{R}_{\imath \kappa}(n)$ &
$\mathbf{R}_{r \jmath}(n)+\mathbf{R}_{\jmath r}(n)+\mathbf{R}_{\kappa \imath}(n)+\mathbf{R}_{\imath \kappa}(n)$ \\
$\Im_\kappa\{\cdot\}$ & $\mathbf{R}_{\kappa
r}(n)+\mathbf{R}_{r
\kappa}(n)-\mathbf{R}_{\imath
\jmath}(n)-\mathbf{R}_{\jmath \imath(n)}$ &
$\mathbf{R}_{\kappa r}(n)-\mathbf{R}_{r
\kappa}(n)+\mathbf{R}_{\imath
\jmath}(n)-\mathbf{R}_{\jmath \imath}(n)$\\
\hline
\end{tabular}
\end{center}
\end{table}
%
\newpage
\noindent Similar duality expressions can be found for $\mathbf{R}_c(n)$, $\mathbf{R}_{\imath}(n)$, $\mathbf{R}_{\jmath}(n)$ and $\mathbf{R}_{\kappa}(n)$ in Table \ref{tab: structure covariance1}-\ref{tab: structure covariance2}. Based on these duality expressions, complete second order statistics in $\mathbb{H}$ are achieved, since all real-valued auto-correlation and cross-correlation matrices can be extracted from the quaternion-valued autocorrelations as
{\small
\!\!\!\!\!\!\!\!\!\!\!\!\begin{align}
\mathbf{R}_{rr
}(n)=\frac{1}{4}\Re\{\mathbf{R}_c(n)+
\mathbf{R}_{\imath}(n)+ \mathbf{R}_{\jmath}(n)+
\mathbf{R}_{\kappa}(n)\},~~
&\mathbf{R}_{\imath\imath}(n)=\frac{1}{4}\Re\{\mathbf{R}_c(n)+
\mathbf{R}_{\imath}(n)- \mathbf{R}_{\jmath}(n)-
\mathbf{R}_{\kappa}(n)
\} \nonumber\\
\mathbf{R}_{\jmath\jmath}(n)=\frac{1}{4}\Re\{\mathbf{R}_c(n)-
\mathbf{R}_{\imath}(n)+ \mathbf{R}_{\jmath}(n)-
\mathbf{R}_{\kappa}(n) \},~~
&\mathbf{R}_{\kappa\kappa}(n)=\frac{1}{4}\Re\{\mathbf{R}_c(n)-
\mathbf{R}_\imath(n)- \mathbf{R}_\jmath (n)+
\mathbf{R}_\kappa (n)
\} \nonumber\\
%
\mathbf{R}_{\imath r}(n)=\frac{1}{4}\Im_\imath\{\mathbf{R}_c(n)
+\mathbf{R}_\imath(n)+\mathbf{R}_{\jmath}(n)+\mathbf{R}_\kappa (n)\},~~
&\mathbf{R}_{\jmath r}(n)=\frac{1}{4}\Im_\jmath\{\mathbf{R}_c(n)
+\mathbf{R}_\imath (n)+\mathbf{R}_{\jmath}(n)+\mathbf{R}_\kappa (n)\}\nonumber\\
\mathbf{R}_{\kappa r}(n)=\frac{1}{4}\Im_\kappa\{\mathbf{R}_c(n)
+\mathbf{R}_\imath (n)+\mathbf{R}_{\jmath}(n)+\mathbf{R}_\kappa (n)\},~~
&\mathbf{R}_{\jmath\imath}(n)=\frac{1}{4}\Im_\kappa\{\mathbf{R}_c(n)
+\mathbf{R}_\imath (n)-\mathbf{R}_{\jmath}(n)-\mathbf{R}_\kappa (n)\}\nonumber\\
\mathbf{R}_{\kappa\imath}(n)=-\frac{1}{4}\Im_\jmath\{\mathbf{R}_c(n)
+\mathbf{R}_\imath (n)-\mathbf{R}_{\jmath}(n)-\mathbf{R}_\kappa (n)\},~~
&\mathbf{R}_{\kappa\jmath}(n)=\frac{1}{4}\Im_\imath\{\mathbf{R}_c(n)
-\mathbf{R}_\imath (n)+\mathbf{R}_{\jmath}(n)-\mathbf{R}_\kappa (n)\}\nonumber\\
\label{eq: complete description}
\end{align}
}
\begin{example}
This numerical example shows the duality between $\mathbb{R}$ and $\mathbb{H}$ in (\ref{eq: complete description}). In other words, the real-valued correlation matrices can be extracted from their quaternion-valued counterparts. Consider the sequence in (\ref{eq: example quaternion sequence}). This sequence can be decomposed into its quadrivariate sequences as follows:
\end{example}
\begin{align}
\mathbf{q}_r=\big[-1,~-2,~-4 \big],~&\mathbf{q}_\imath=\big[-10,~-4,~-5 \big]\nonumber\\
\mathbf{q}_\jmath =\big[1,~-6,~3 \big],~&\mathbf{q}_\kappa =\big[-1,~3,~1 \big]
\label{eq: example 2 cct}
\end{align}
%
%
Their autocorrelation sequences can be straightforwardly calculated as follows
\begin{align}
   \mathbf{r}_{rr} = \big[1.33,~3.33,~\underset{\mathlarger{\mathbf{\uparrow}}}{7},~3.33,~1.33\big],~
   & \mathbf{r}_{\imath \imath} =\big[ 16.67,~20,~\underset{\mathlarger{\mathbf{\uparrow}}}{47},~20,~16.67\big]\nonumber\\ 
   \mathbf{r}_{\jmath \jmath} = \big[1,~-8,~ \underset{\mathlarger{\mathbf{\uparrow}}}{15.33},~-8,~1\big],~ 
   & \mathbf{r}_{\kappa \kappa} =\big[-0.33,~0,~\underset{\mathlarger{\mathbf{\uparrow}}}{3.67},~0,~-0.33\big]\nonumber\\
   \mathbf{r}_{\imath r} = \big[13.33,~12,~\underset{\mathlarger{\mathbf{\uparrow}}}{12.67},~4.67,~1.67\big],~
   &\mathbf{r}_{\jmath r} =\big[-1.33,~7.33,~\underset{\mathlarger{\mathbf{\uparrow}}}{-0.33},~0,~-1\big]\nonumber\\
   \mathbf{r}_{kr} =\big[ 1.33,~-3.33,~\underset{\mathlarger{\mathbf{\uparrow}}}{-3},~-1.67,~-0.33\big],~
   &\mathbf{r}_{\jmath \imath} = \big[-1.67,~8.67,~\underset{\mathlarger{\mathbf{\uparrow}}}{-0.33},~16,~-10\big]\nonumber\\
   \mathbf{r}_{\kappa \imath} =\big[1.67,~-3.67,~\underset{\mathlarger{\mathbf{\uparrow}}}{-2.33},-11.33,~-3.33\big],~
   & \mathbf{r}_{\kappa \jmath} =\big[-1,~5,~\underset{\mathlarger{\mathbf{\uparrow}}}{-5.33},~-1,~0.33\big]\nonumber\\\label{eq:autocorrelation real}
\end{align}
Recall that the arrow symbol $\uparrow$ denotes the lag $\ell=0$. For example, $r_{\imath r}(0)=12.67$ shows the cross-correlation between the $\imath-$imaginary and real component of (\ref{eq: example quaternion sequence}) at zero lag, while $r_{\imath r}(-1)=12$ corresponds to lag $\ell=-1$. These real-valued autocorrelation sequences can be used to construct their real-valued autocorrelation matrices shown in Table \ref{tab: autocorrelation real matrices}. These real-valued autocorrelation matrices can also be extracted from their quaternion counterparts in Table \ref{tab: autocorrelation quaternion matrices} based on the duality expressions in (\ref{eq: complete description}). This shows how the augmented statistics of quaternions in (\ref{eq: autocorrelation c})-(\ref{eq: autocorrelation ijk}) can provide the complete second order statistics in the context of autocorrelation and cross-correlation functions.
\\\indent The same principle can be extended to other statistical descriptors, such as covariances, moments, and cumulants. Now that it has been shown how the augmented basis leads to the complete statistics of quaternions, the next section shows how this augmented basis can be put in practice in quaternion signal processing.
\begin{table}
\caption{Real-valued autocorrelation matrices corresponding to their quaternion-valued counterparts in Table \ref{tab: autocorrelation quaternion matrices}. The double subscripts indicate the correlation between the two relevant quaternion components. For example, $R_{\imath r}(n)$ corresponds to the cross correlation between the $\imath-$imaginary and real component.}\label{tab: autocorrelation real matrices}
 \begin{tabular}{|*2{>{\centering\arraybackslash}p{0.5\textwidth}|}}
\hline
  \multicolumn{2}{|c|}{Autocorrelation matrices in $\mathbb{R}$ (of size $3 \times 3$)}  \\
\hline
\[\mathbf{R}_{rr
}(n)=\left[ \begin{array}{ccc} 7     & 3.33 & 1.33\\ 
                                3.33 & 7 & 3.33\\
                                1.33& 3.33 & 7\end{array}\right]\] 
& \[\mathbf{R}_{\imath\imath}(n)
= \left[ \begin{array}{ccc} 47 & 20 & 16.67\\ 
                             20 & 47 & 20 \\
                              16.67  & 20 & 47 \end{array}\right]\]
\\\hline
\[\mathbf{R}_{\jmath\jmath
}(n)=\left[ \begin{array}{ccc} 15.33 & -8 & 1\\ 
                                -8 & 15.33 & -8\\
                               1 & -8 & 15.33\end{array}\right]\] 
& \[\mathbf{R}_{\kappa\kappa}(n)
        = \left[ \begin{array}{ccc} 3.67 & 0 & -0.33\\                                                    0 & 3.67  &0\\
                                -0.33 &0 & 3.67\end{array}\right]\]
\\\hline
\[\mathbf{R}_{\imath r
}(n)=\left[ \begin{array}{ccc} 12.67 & 4.67 & 1.67\\ 
                                12 & 12.67 & 4.67\\
                                13.33& 12 & 12.67\end{array}\right]\] 
& \[\mathbf{R}_{\jmath r}(n) 
                    = \left[ \begin{array}{ccc} -0.33 & 0&-1 \\ 
                                                7.33 & -0.33&0\\
                                                -1.33&7.33 &-0.33\end{array}\right]\]
\\\hline
\[\mathbf{R}_{\kappa r
}(n)=\left[ \begin{array}{ccc} -3 & -1.67 & -0.33\\ 
                                -3.33 & -3 & -1.67\\
                                1.33&-3.33&-3 \end{array}\right]\] 
& \[\mathbf{R}_{\jmath\imath}(n)
                = \left[ \begin{array}{ccc} -0.33 & 16 &-10\\ 
                                            8.67 & -0.33 & 16\\
                                        -1.67&8.67&-0.33\end{array}\right]\]
\\\hline
\[\mathbf{R}_{\kappa \imath
}(n)=\left[ \begin{array}{ccc} -2.33 & -11.33 & -3.33\\ 
                                -3.67 & -2.33 & -11.33\\
                                1.67& -3.67 & -2.33\end{array}\right]\] 
& \[\mathbf{R}_{\kappa\jmath}(n) 
= \left[ \begin{array}{ccc} -5.33 & -1 &0.33\\
                            5 & -5.33 &-1 \\
                            -1& 5 & -5.33\end{array}\right]\]
\\\hline
\end{tabular}
\end{table}
%
\section{MMSE Estimation and Widely linear Model}

\label{Sec:MMSE}

In order to better emphasise the importance of the augmented approach, consider the MMSE estimator of a variable, $y$, conditioned on the observation, $z$, that is $\hat{y}= \E{y|z}$. Now, assume that $y$ and $z$ are quaternion-valued zero-mean and jointly Gaussian random variables.\footnote{A quaternion-valued random variable is Gaussian if its real-valued components are jointly Gaussian, while set of quaternion-valued random variables are jointly Gaussian if their real-valued components are jointly Gaussian~\cite{AQS}.} In this case, the MMSE estimator has to be expressed according to the individual components of the quaternion random variables, that is 
\[
\begin{aligned}
\hat{y} =&\E{y_{r}|z_{r},z_{\imath}, z_{
\jmath},z_{\kappa}} + \imath\E{y_{\imath}|z_{r},z_{\imath},z_{\jmath},z_{\kappa}} + 
\\
&+\jmath\E{y_{\jmath}|z_{r},z_{\imath},z_{\jmath}, z_{\kappa}} + \kappa\E{y_{\kappa}|z_{r},z_{\imath},z_{\jmath},z_{\kappa}}
\end{aligned}
\]
At this point we can use the mapping derived in (\ref{eq:MappingHR}) to replace the quaternion components $\{z_{r}$,$z_{\imath}$,$z_{\jmath}$,$z_{\kappa}\}$ with quaternion involutions $\{z,z^{\imath},z^{\jmath},z^{\kappa}\}$ yielding
\[
\begin{aligned}
\hat{y} =&\E{y_{r}|z,z^{\imath}, z^{\jmath},z^{\kappa}} + \imath\E{y_{\imath}|z,z^{\imath},z^{\jmath},z^{\kappa}}
\\
&+\jmath\E{y_{\jmath}|z,z^{\imath},z^{\jmath},z^{\kappa}} + \kappa\E{y_{\kappa}|z,z^{\imath},z^{\jmath},z^{\kappa}}
\end{aligned}
\]
Therefore, for quaternion-valued, zero-mean, and jointly Gaussian $z$ and $y$, the MMSE solution is given by the widely linear model~\cite{AQS} 
\begin{equation}
\hat{y} = \mathbf{g}^{\T}\mathbf{z}+\mathbf{h}^{\T}\mathbf{z}^{\imath} + \mathbf{u}^{\T}\mathbf{z}^{\jmath}+\mathbf{v}^{\T}\mathbf{z}^{\kappa}=\begin{bmatrix}
\mathbf{g}^{\T},\mathbf{h}^{\T},\mathbf{u}^{\T},\mathbf{v}^{\T}\end{bmatrix}\mathbf{z}^{a}
\label{eq:widely-linear-estimator}
\end{equation}
where $\{\mathbf{g},\mathbf{h},\mathbf{u},\mathbf{v}\}$ are quaternion-valued coefficient vectors and $\mathbf{z}$ is the regressor vector. The estimator in (\ref{eq:widely-linear-estimator}) can now be formulated into 
\begin{equation}
\hat{\mathbf{y}}^{a}=\begin{bmatrix}\mathbf{g}^{\T}&\mathbf{h}^{\T}&\mathbf{u}^{\T}&\mathbf{v}^{\T}\\\mathbf{h}^{\imath\T}&\mathbf{g}^{\imath\T}&\mathbf{v}^{\imath\T}&\mathbf{u}^{\imath\T}\\\mathbf{u}^{\jmath\T}&\mathbf{v}^{\jmath\T}&\mathbf{g}^{\jmath\T}&\mathbf{h}^{\jmath\T}\\\mathbf{v}^{\kappa\T}&\mathbf{u}^{\kappa\T}&\mathbf{h}^{\kappa\T}&\mathbf{g}^{\kappa\T}\end{bmatrix}\mathbf{z}^{a}
\label{eq:augmented-estimator}
\end{equation}
As the input, $z$, must be used in its augmented representation, the approach is referred to as augmented. In addition, since the estimators in \eqref{eq:widely-linear-estimator} and \eqref{eq:augmented-estimator} are linear with respect to the augmented input vector, they are also known as widely linear estimators, denoting linearity of the models in a wider sense. To optimise the weights in the widely linear model (\ref{eq:widely-linear-estimator}), quaternion derivatives are required for signal processing algorithms. Thus, the quaternion calculus is next introduced.


\section{Quaternion Calculus}

\label{Sec:CalQ}

In general, the traditional definition of derivative is too restrictive in the quaternion domain, since they accommodate only linear quaternion-valued functions~\cite{GenHR}. Thus, there have been efforts to relax these conditions. One notable effort in this front  has been to extend the framework of analyticity from the complex domain to quaternion-valued functions and introduce an equivalent condition to the Cauchy-Riemann condition for analyticity of complex-valued functions. These efforts have resulted in the Cauchy-Riemann-Fueter condition~\cite{FueterGerman,GenHR}\footnote{Due to non-commutativity of quaternion algebra, an alternative formulation can be given for the Cauchy–Riemann–Fueter condition, that is~\cite{HR-Clive,HR-Gradient,GenHR} 
\[
\frac{\partial \mathrm{f}}{\partial \mathbf{q}_{r}} + \frac{\partial \mathrm{f}}{\partial \mathbf{q}_{\imath}}\imath + \frac{\partial \mathrm{f}}{\partial \mathbf{q}_{\jmath}}\jmath + \frac{\partial \mathrm{f}}{\partial \mathbf{q}_{\kappa}}\kappa = 0
\]
The only difference between the two derivations of the Cauchy–Riemann–Fueter condition is that one is derived based on placement of $\imath$, $\jmath$, and $\kappa$ on the left-hand-side of the derivatives and the other is derived based on placement of these imaginary units on the right-hand-side of the derivatives.}
\begin{equation}
\frac{\partial \mathrm{f}}{\partial \mathbf{q}_{r}} + \imath\frac{\partial \mathrm{f}}{\partial \mathbf{q}_{\imath}} + \jmath\frac{\partial \mathrm{f}}{\partial \mathbf{q}_{\jmath}} + \kappa\frac{\partial \mathrm{f}}{\partial \mathbf{q}_{\kappa}} = 0
\label{eq:CRF}
\end{equation} 
for quaternion-valued function $\mathrm{f}\left(\cdot\right):\mathbb{H}^{m}\rightarrow\mathbb{H}$. The Cauchy-Riemann-Fueter condition, however, has also proven to be overly restrictive for use as the basis for deriving quaternion-valued optimisation, adaptation, and other information processing machinery. This issue has rekindled efforts to relax the condition in \eqref{eq:CRF} even further. Important results in this arena include the notion of local analyticity~\cite{Leo}. In this setting, the function $\mathrm{f}\left(\cdot\right):\mathbb{H}\rightarrow\mathbb{H}$
behaves akin to an analytical function at $q$ if
\begin{equation}
\frac{\partial\mathrm{f}\left(q\right)}{\partial q_r} = -\xi\frac{\partial\mathrm{f}\left(q\right)}{\partial\|\Im\{q\}\|}\hspace{0.32cm}\text{with}\hspace{0.32cm}\xi=\frac{\Im\{q\}}{\|\Im\{q\}\|}
\label{eq:local_analyticity}
\end{equation}
i.e. the quaternion behaves as a complex number. An important result from local analyticity is the derivative of the exponential function~\cite{QNAF}
\begin{equation}
\frac{\partial e^{q}}{\partial q}= e^{q}
\end{equation}
Despite numerous efforts to relax differentiability conditions for quaternion-valued functions, these conditions remain too restrictive. Indeed, this issue has been one of the main stumbling block stiffening the derivation of quaternion-valued signal processing and learning techniques. Although the conditions for quaternion analyticity are restrictive, it is not always required to work with quaternion analytical functions. For the major part, we only need to establish the rate of change for a function in different directions and/or calculate the direction of maximum rate of change. An elegant solution to this problem has been the establishment of the $\mathbb{HR}$-calculus~\cite{GenHR}. 

Consider the quaternion-valued function $\mathrm{f}\left(\cdot\right):\mathbb{H}^{m}\rightarrow\mathbb{H}$, so that we can write 
\begin{equation}
\begin{aligned}
\mathrm{f}\left(\mathbf{q}\right)=
\mathrm{f}\left(\mathbf{q}_{r},\mathbf{q}_{\imath},\mathbf{q}_{\jmath},\mathbf{q}_{\kappa}\right)=&\mathrm{f}_{r}\left(\mathbf{q}_{r},\mathbf{q}_{\imath},\mathbf{q}_{\jmath},\mathbf{q}_{\kappa}\right)+\imath\mathrm{f}_{\imath}\left(\mathbf{q}_{r},\mathbf{q}_{\imath},\mathbf{q}_{\jmath},\mathbf{q}_{\kappa}\right)
\\
&+\jmath\mathrm{f}_{\jmath}\left(\mathbf{q}_{r},\mathbf{q}_{\imath},\mathbf{q}_{\jmath},\mathbf{q}_{\kappa}\right)+\kappa\mathrm{f}_{\kappa}\left(\mathbf{q}_{r},\mathbf{q}_{\imath},\mathbf{q}_{\jmath},\mathbf{q}_{\kappa}\right)
\end{aligned}
\end{equation}
where $\{\mathrm{f}_{r}\left(\mathbf{q}\right),\mathrm{f}_{\imath}\left(\mathbf{q}\right),\mathrm{f}_{\jmath}\left(\mathbf{q}\right),\mathrm{f}_{\kappa}\left(\mathbf{q}\right)\}$ denote real-valued components comprising $\mathrm{f}\left(\mathbf{q}\right)$ with each being real-valued functions of real-valued vectors $\{\mathbf{q}_{r},\mathbf{q}_{\imath},\mathbf{q}_{\jmath},\mathbf{q}_{\kappa}\}$. The $\mathbb{HR}$-calculus, on one hand, considers the function $\mathrm{f}\left(\mathbf{q}\right)$ as
\begin{equation}
\mathrm{f}\left(\mathbf{q}\right)=\mathrm{f}\left(\mathbf{q}_{r}+\imath\mathbf{q}_{\imath}+\jmath\mathbf{q}_{\jmath}+\kappa\mathbf{q}_{\kappa}\right)=\mathrm{f}\left(\mathbf{q}^{a}=\begin{bmatrix}\mathbf{q}^{\T}&\mathbf{q}^{\imath\T}&\mathbf{q}^{\jmath\T}&\mathbf{q}^{\kappa\T}\end{bmatrix}^{\T}\right)
\end{equation}
where \eqref{eq:MappingHRinRev} is used to replace $\{\mathbf{q}_{r},\mathbf{q}_{\imath},\mathbf{q}_{\jmath},\mathbf{q}_{\kappa}\}$ with $\mathbf{q}^{a}$. On the other hand, the $\mathbb{HR}$-calculus considers formulation of $\mathrm{f}\left(\cdot\right):\mathbb{H}^{m}\rightarrow\mathbb{H}$ as the quadrivariate function formulated as
\begin{equation}
\mathrm{g}\left(\mathbf{q}_{r},\mathbf{q}_{\imath},\mathbf{q}_{\jmath},\mathbf{q}_{\kappa}\right)=\begin{bmatrix}\mathrm{f}_{r}\left(\mathbf{q}_{r},\mathbf{q}_{\imath},\mathbf{q}_{\jmath},\mathbf{q}_{\kappa}\right)
\\
\imath\mathrm{f}_{\imath}\left(\mathbf{q}_{r},\mathbf{q}_{\imath},\mathbf{q}_{\jmath},\mathbf{q}_{\kappa}\right)
\\
\jmath\mathrm{f}_{\jmath}\left(\mathbf{q}_{r},\mathbf{q}_{\imath},\mathbf{q}_{\jmath},\mathbf{q}_{\kappa}\right)
\\
\kappa\mathrm{f}_{\kappa}\left(\mathbf{q}_{r},\mathbf{q}_{\imath},\mathbf{q}_{\jmath},\mathbf{q}_{\kappa}\right)\end{bmatrix}
\label{eq:DefG}
\end{equation}
The gradient of the function formulated in \eqref{eq:DefG} can be calculated using classical vector calculus as
\begin{equation}
\nabla\mathrm{g}=\begin{bmatrix}\frac{\partial \mathrm{g} (\mathbf{q}_{r},\mathbf{q}_{\imath},\mathbf{q}_{\jmath},\mathbf{q}_{\kappa})}{\partial \mathbf{q}_{r}}\\
\frac{\partial \mathrm{g} (\mathbf{q}_{r},\mathbf{q}_{\imath},\mathbf{q}_{\jmath},\mathbf{q}_{k})}{\partial \mathbf{q}_{\imath}}\\
\frac{\partial \mathrm{g} (\mathbf{q}_{r},\mathbf{q}_{\imath},\mathbf{q}_{\jmath},\mathbf{q}_{\kappa})}{\partial \mathbf{q}_{\jmath}}\\
\frac{\partial \mathrm{g} (\mathbf{q}_{r},\mathbf{q}_{\imath},\mathbf{q}_{\jmath},\mathbf{q}_{\kappa})}{\partial \mathbf{q}_{\kappa}}
\end{bmatrix} 
\end{equation}
The $\mathbb{HR}$-calculus exploits the expressions in \eqref{eq:RealValuedComponents} to formulate the gradient of $\mathrm{g}\left(\mathbf{q}_{r},\mathbf{q}_{\imath},\mathbf{q}_{\jmath},\mathbf{q}_{\kappa}\right)$ in terms of the augmented quaternion vector $\mathbf{q}^{a}$ and $\mathrm{f}\left(\mathbf{q}^{a}\right)$, resulting in\footnote{The blue font highlights the derivative of interest, i.e. derivative with respect to the conjugate quaternion, which is sufficient for most algorithm derivations.}
\begin{equation}
\nabla_{\mathbf{q}^{a*}}\mathrm{f}\left(\mathbf{q}\right)=
\begin{bmatrix}{\color{blue}
\frac{\partial \mathrm{f} (\mathbf{q},\mathbf{q}^{\imath},\mathbf{q}^{\jmath},\mathbf{q}^{\kappa})}{\partial \mathbf{q}^{*}}}\\
\frac{\partial \mathrm{f} (\mathbf{q},\mathbf{q}^{\imath},\mathbf{q}^{\jmath},\mathbf{q}^{\kappa})}{\partial \mathbf{q}^{\imath*}}\\
\frac{\partial \mathrm{f} (\mathbf{q},\mathbf{q}^{\imath},\mathbf{q}^{\jmath},\mathbf{q}^{\kappa})}{\partial \mathbf{q}^{\jmath*}}\\
\frac{\partial \mathrm{f} (\mathbf{q},\mathbf{q}^{\imath},\mathbf{q}^{\jmath},\mathbf{q}^{\kappa})}{\partial \mathbf{q}^{\kappa*}}
\end{bmatrix}
=\frac{1}{4}\underbrace{\begin{bmatrix}{\color{blue}\mathbf{I}}&{\color{blue}\imath\mathbf{I}}&{\color{blue}\jmath\mathbf{I}}&{\color{blue}\kappa\mathbf{I}}\\\mathbf{I}&\imath\mathbf{I}&-\jmath\mathbf{I}&-\kappa\mathbf{I}\\\mathbf{I}&-\imath\mathbf{I}&\jmath\mathbf{I}&-\kappa\mathbf{I}\\\mathbf{I}&-\imath\mathbf{I}&-\jmath\mathbf{I}&\kappa\mathbf{I}\end{bmatrix}}_{\mathbf{A}}\underbrace{\begin{bmatrix}{\color{blue}\frac{\partial \mathrm{g} (\mathbf{q}_{r},\mathbf{q}_{\imath},\mathbf{q}_{\jmath},\mathbf{q}_{\kappa})}{\partial \mathbf{q}_{r}}}\\
{\color{blue}\frac{\partial \mathrm{g} (\mathbf{q}_{r},\mathbf{q}_{\imath},\mathbf{q}_{\jmath},\mathbf{q}_{k})}{\partial \mathbf{q}_{\imath}}}\\
{\color{blue}\frac{\partial \mathrm{g} (\mathbf{q}_{r},\mathbf{q}_{\imath},\mathbf{q}_{\jmath},\mathbf{q}_{\kappa})}{\partial \mathbf{q}_{\jmath}}}\\{\color{blue}
\frac{\partial \mathrm{g} (\mathbf{q}_{r},\mathbf{q}_{\imath},\mathbf{q}_{\jmath},\mathbf{q}_{\kappa})}{\partial \mathbf{q}_{\kappa}}}
\end{bmatrix}}_{\nabla\mathrm{g}}
\label{eq:QGO}
\end{equation}
and the direction of maximum change given by
\begin{equation}
\boxed{\frac{\partial \mathrm{f} \left(\mathbf{q}^{a}\right)}{\partial\mathbf{q}^{*}}=\frac{1}{4}\left(\frac{\partial\mathrm{f}\left(\mathbf{q}^{a}\right)}{\partial\mathbf{q}_{r}}+\imath\frac{\partial\mathrm{f}\left(\mathbf{q}^{a}\right)}{\partial\mathbf{q}_{\imath}}+\jmath\frac{\partial\mathrm{f}\left(\mathbf{q}^{a}\right)}{\partial\mathbf{q}_{\jmath}}+\kappa\frac{\partial \mathrm{f}\left(\mathbf{q}^{a}\right)}{\partial\mathbf{q}_{\kappa}}\right)
}
\label{eq:grad}
\end{equation}
with the full derivation for \eqref{eq:QGO} and \eqref{eq:grad} given in~\cite{PouriaPhD}. 
An alternative definition can also be formulated as 
\begin{equation}
\frac{\partial_{\text{left}} \mathrm{f} \left(\mathbf{q}^{a}\right)}{\partial\mathbf{q}^{*}}=\frac{1}{4}\left(\frac{\partial\mathrm{f}\left(\mathbf{q}^{a}\right)}{\partial\mathbf{q}_{r}}+\frac{\partial\mathrm{f}\left(\mathbf{q}^{a}\right)}{\partial\mathbf{q}_{\imath}}\imath+\frac{\partial\mathrm{f}\left(\mathbf{q}^{a}\right)}{\partial\mathbf{q}_{\jmath}}\jmath+\frac{\partial\mathrm{f}\left(\mathbf{q}^{a}\right)}{\partial\mathbf{q}_{\kappa}}\kappa\right)
\label{eq:grad-left}
\end{equation}
which is denoted as left derivatives due to placement of the partial derivatives to the left-hand-side of the imaginary units. Although both the right-hand-side and left-hand-side derivatives yield equivalent results, a skilful use of both approaches can simplify the derived algorithms. In the remainder of this section we will; i) introduce derivatives of functions useful as base cases, ii) present derivative rules to extract differentials of more complex functions.

\begin{example}
Consider the second-order norm function given by
\begin{equation}
\forall q\in\mathbb{H}:\hspace{0.32cm} \mathrm{f}\left(q\right)=qq^{*}=\|q\|^{2}=q^{2}_{r}+q^{2}_{\imath}+q^{2}_{\jmath}+q^{2}_{\kappa}
\end{equation}
The second-order norm function appears as the basis for cost functions in a large number of adaptation and learning techniques. The augmented and quadrivariate formulations of the second-order norm are given by
\begin{equation}
\mathbf{f}\left(\mathbf{q}^{a}\right)=\frac{1}{4}\mathbf{q}^{a\H}\mathbf{q}^{a}\hspace{0.32cm}\text{and}\hspace{0.32cm}\mathrm{g}\left(q_{r},q_{\imath},q_{\jmath},q_{\kappa}\right)=\begin{bmatrix}q^{2}_{r}+q^{2}_{\imath}+q^{2}_{\jmath}+q^{2}_{\kappa}\\0\\0\\0\end{bmatrix}
\end{equation}
From \eqref{eq:QGO}, it now follows that the gradient of the second-order norm is
\begin{equation}
\nabla_{\mathbf{q}^{a*}}\mathrm{f}\left(\mathbf{q}^{a}\right)
=\frac{1}{4}\begin{bmatrix}
1&\phantom{-}\imath&\phantom{-}\jmath&\phantom{-}\kappa\\
1&\phantom{-}\imath&-\jmath&-\kappa\\
1&-\imath&\phantom{-}\jmath&-\kappa\\
1&-\imath&-\jmath&\phantom{-}\kappa
\end{bmatrix}
\begin{bmatrix}
2q_{r}\\2q_{\imath}\\2q_{\jmath}\\2q_{\kappa}
\end{bmatrix}=\frac{1}{2}\begin{bmatrix}
q\\q^{\imath}\\q^{\jmath}\\q^{\kappa}
\end{bmatrix}
\end{equation}
establishing that 
\begin{equation}
\frac{\partial\|q\|^{2}}{\partial q^{*}}=\frac{1}{2}q
\end{equation}
which is different from results in the $\mathbb{CR}$-calculus. 
\end{example}

\begin{example}
Now, let us move to norm operator
\begin{equation}
\forall q\in\mathbb{H}:\hspace{0.32cm} \mathrm{f}\left(q\right)=\sqrt{qq^{*}}=\|q\|=\sqrt{q^{2}_{r}+q^{2}_{\imath}+q^{2}_{\jmath}+q^{2}_{\kappa}}
\end{equation}
In this case, the augmented and quadrivariate formulations of the second-order norm are given by
\begin{equation}
\mathbf{f}\left(\mathbf{q}^{a}\right)=\frac{1}{2}\sqrt{\mathbf{q}^{a\H}\mathbf{q}^{a}}\hspace{0.32cm}\text{and}\hspace{0.32cm}\mathrm{g}\left(q_{r},q_{\imath},q_{\jmath},q_{\kappa}\right)=\begin{bmatrix}\sqrt{q^{2}_{r}+q^{2}_{\imath}+q^{2}_{\jmath}+q^{2}_{\kappa}}\\0\\0\\0\end{bmatrix}
\end{equation}
Using the relation in \eqref{eq:QGO}, we can now write
\begin{equation}
\nabla_{\mathbf{q}^{a*}}\mathrm{f}\left(\mathbf{q}^{a}\right)
=\frac{1}{4}\begin{bmatrix}
1&\phantom{-}\imath&\phantom{-}\jmath&\phantom{-}\kappa\\
1&\phantom{-}\imath&-\jmath&-\kappa\\
1&-\imath&\phantom{-}\jmath&-\kappa\\
1&-\imath&-\jmath&\phantom{-}\kappa
\end{bmatrix}
\begin{bmatrix}
\frac{q_{r}}{\sqrt{q^{2}_{r}+q^{2}_{\imath}+q^{2}_{\jmath}+q^{2}_{\kappa}}}\\\frac{q_{\imath}}{\sqrt{q^{2}_{r}+q^{2}_{\imath}+q^{2}_{\jmath}+q^{2}_{\kappa}}}\\\frac{q_{\jmath}}{\sqrt{q^{2}_{r}+q^{2}_{\imath}+q^{2}_{\jmath}+q^{2}_{\kappa}}}\\\frac{q_{\kappa}}{\sqrt{q^{2}_{r}+q^{2}_{\imath}+q^{2}_{\jmath}+q^{2}_{\kappa}}}
\end{bmatrix}=\frac{1}{4}\begin{bmatrix}
\frac{q}{\|q\|}\\\frac{q^{\imath}}{\|q\|}\\\frac{q^{\jmath}}{\|q\|}\\\frac{q^{\kappa}}{\|q\|}
\end{bmatrix}
\end{equation}
Therefore, establishing
\begin{equation}
\frac{\partial\|q\|}{\partial q^{*}}=\frac{1}{4}\frac{q}{\|q\|}
\end{equation}

\end{example}

\begin{example}
Let us now consider $\mathrm{f}\left(q\right)=q^{2}$. 
In this case, we have
\begin{equation}
\mathrm{g}\left(q_{r},q_{\imath},q_{\jmath},q_{\kappa}\right)=\begin{bmatrix}q^{2}_{r}-q^{2}_{\imath}-q^{2}_{\jmath}-q^{2}_{\kappa}\\\imath2q_{r}q_{\imath}\\\jmath2q_{r}q_{\jmath}\\\kappa2q_{r}q_{\kappa}\end{bmatrix}\hspace{0.32cm}\text{and}\hspace{0.32cm}\nabla\mathrm{g}=\begin{bmatrix}2q_{r}+\imath2q_{\imath}+\jmath2q_{\jmath}+2\kappa q_{\kappa}\\\imath2q_{r}-2q_{\imath}\\\jmath2q_{r}-2q_{\jmath}\\\kappa2q_{r}-2q_{\kappa}\end{bmatrix}
\end{equation}
Therefore, using \eqref{eq:QGO} results in 
\begin{equation}
\nabla_{\mathbf{q}^{a*}}\left(\mathbf{q}^{2}\right)=
\begin{bmatrix}
\frac{\partial \mathrm{f} (q^{*},q^{\imath*},q^{\jmath*},q^{\kappa*})}{\partial q^{*}}\\
\frac{\partial \mathrm{f}(q^{*},q^{\imath*},q^{\jmath*},q^{\kappa*})}{\partial q^{\imath*}}\\
\frac{\partial \mathrm{f} (q^{*},q^{\imath*},q^{\jmath*},q^{\kappa*})}{\partial q^{\jmath*}}\\
\frac{\partial \mathrm{f} (q^{*},q^{\imath*},q^{\jmath*},q^{\kappa*})}{\partial q^{\kappa*}}
\end{bmatrix}
=\begin{bmatrix}-\Re\{q\}\\q-\Im_{\imath}\{q\}\\q-\Im_{\jmath}\{q\}\\q-\Im_{\kappa}\{q\}\end{bmatrix}
\label{eq:Diffq2}
\end{equation}
where
\begin{equation}
\frac{\partial \mathrm{f}\left(q,q^{\imath},q^{\jmath},q^{\kappa}\right)}{\partial q^{*}}=\frac{1}{4}\left(\frac{\partial \mathrm{f}}{\partial q_{r}} + \imath\frac{\partial \mathrm{f}}{\partial q_{\imath}} + \jmath\frac{\partial \mathrm{f}}{\partial q_{\jmath}} + \kappa\frac{\partial \mathrm{f}}{\partial q_{\kappa}}\right)=-\Re\{q\}
\end{equation}

\end{example}

\begin{example}
Consider the quaternion rectified linear unit (QReLU) given by 
\begin{equation}
\text{QReLU}\left(q\right)=\text{ReLU}\left(q_{r}\right)+\imath\text{ReLU}\left(q_{\imath}\right)+\jmath\text{ReLU}\left(q_{\jmath}\right)+\kappa\text{ReLU}\left(q_{\kappa}\right)
\end{equation}
where $\text{ReLU}\left(\cdot\right)$ denotes the real-valued ReLU function. In this case, we have
\begin{equation}
\mathrm{g}\left(q_{r},q_{\imath},q_{\jmath},q_{\kappa}\right)=\begin{bmatrix}\phantom{l}\text{ReLU}\left(q_{r}\right)\\\imath\text{ReLU}\left(q_{\imath}\right)\\\jmath\text{ReLU}\left(q_{\jmath}\right)\\\kappa\text{ReLU}\left(q_{\kappa}\right)\end{bmatrix}
\end{equation}
that has the gradient
\begin{equation}
\nabla\mathrm{g}\left(q_{r},q_{\imath},q_{\jmath},q_{\kappa}\right)=\begin{bmatrix}\mathrm{u}
\left(q_{r}\right)\\\imath\mathrm{u}
\left(q_{\imath}\right)\\
\jmath\mathrm{u}\left(q_{\jmath}\right)\\
\kappa\mathrm{u}\left(q_{\kappa}\right)
\end{bmatrix}
\end{equation}
where
\begin{equation}
\forall x\in\mathbb{R}:\hspace{0.12cm}\mathrm{u}\left(x\right)=\left\{\begin{matrix}1&\text{if }x>0\\0&\text{otherwise}\end{matrix}\right.
\end{equation}
Now using the relation in \eqref{eq:QGO} yields
\begin{equation}
\nabla_{\mathbf{q}^{a*}}\text{QReLU}\left(q\right)=\frac{1}{4}\begin{bmatrix}
1&\phantom{-}\imath&\phantom{-}\jmath&\phantom{-}\kappa\\
1&\phantom{-}\imath&-\jmath&-\kappa\\
1&-\imath&\phantom{-}\jmath&-\kappa\\
1&-\imath&-\jmath&\phantom{-}\kappa
\end{bmatrix}\begin{bmatrix}\mathrm{u}
\left(q_{r}\right)\\\imath\mathrm{u}
\left(q_{\imath}\right)\\
\jmath\mathrm{u}\left(q_{\jmath}\right)\\
\kappa\mathrm{u}\left(q_{\kappa}\right)
\end{bmatrix}
\end{equation}
and therefore
\begin{equation}
\frac{\partial\text{QReLU}\left(q\right)}{\partial q^{*} }=\frac{1}{4}\left(\mathrm{u}
\left(q_{r}\right)-\mathrm{u}
\left(q_{\imath}\right)-\mathrm{u}\left(q_{\jmath}\right)-\mathrm{u}\left(q_{\kappa}\right)\right)
\end{equation}
\end{example}


\subsection{Multiplication Rules}

Let us begin with multiplication on the left-hand-side by a constant quaternion number, $\nu\in\mathbb{H}$. In this case, from \eqref{eq:grad}, we can write 
\begin{equation}
\begin{aligned}
\frac{\partial\nu\mathrm{f}\left(\mathbf{q}^{a}\right)}{\partial \mathbf{q}^{*}}=&\frac{1}{4}\left(\frac{\partial\nu\mathrm{f}\left(\mathbf{q}^{a}\right)}{\partial\mathbf{q}_{r}}+\imath\frac{\partial\nu\mathrm{f}\left(\mathbf{q}^{a}\right)}{\partial\mathbf{q}_{\imath}}+\jmath\frac{\partial\nu\mathrm{f}}{\partial\mathbf{q}_{\jmath}}+\kappa\frac{\partial\nu \mathrm{f}\left(\mathbf{q}^{a}\right)}{\partial\mathbf{q}_{\kappa}}\right)
\\
=&\frac{1}{4}\left(\nu\frac{\partial\mathrm{f}\left(\mathbf{q}^{a}\right)}{\partial\mathbf{q}_{r}}+\imath\nu\frac{\partial\mathrm{f}\left(\mathbf{q}^{a}\right)}{\partial\mathbf{q}_{\imath}}+\jmath\nu\frac{\partial\mathrm{f}\left(\mathbf{q}^{a}\right)}{\partial\mathbf{q}_{\jmath}}+\kappa\nu\frac{\partial \mathrm{f}\left(\mathbf{q}^{a}\right)}{\partial\mathbf{q}_{\kappa}}\right)
\end{aligned}
\label{eq:PreImportantReal}
\end{equation}
In order to factor out $\nu$ from \eqref{eq:PreImportantReal}, we need to change the order of multiplications involving $\nu$. Focusing on $\imath$ and $\nu$, as an example case, we use $\nu\nu^{-1}=1$, and the definition of quaternion involution in \eqref{eq:InvoQ} to yield 
\begin{equation}
\imath\nu=\nu\nu^{-1}\imath\nu=\nu\imath^{\nu^{-1}}
\label{eq:Change2}
\end{equation}
Similarly, it can be shown that
\begin{equation}
\jmath\nu=\nu\jmath^{\nu^{-1}}\hspace{0.32cm}\text{and}\hspace{0.32cm}\kappa\nu=\nu\kappa^{\nu^{-1}}
\label{eq:Change1}
\end{equation}
Now, replacing \eqref{eq:Change2} and \eqref{eq:Change1} into \eqref{eq:PreImportantReal} to factor out ``$\nu$'' yields 
\begin{equation}
\begin{aligned}
\frac{\partial\nu\mathrm{f}\left(\mathbf{q}^{a}\right)}{\partial \mathbf{q}^{*}}=&\frac{1}{4}\left(\nu\frac{\partial\mathrm{f}\left(\mathbf{q}^{a}\right)}{\partial\mathbf{q}_{r}}+\imath\nu\frac{\partial\mathrm{f}\left(\mathbf{q}^{a}\right)}{\partial\mathbf{q}_{\imath}}+\jmath\nu\frac{\partial\mathrm{f}\left(\mathbf{q}^{a}\right)}{\partial\mathbf{q}_{\jmath}}+\kappa\nu\frac{\partial \mathrm{f}\left(\mathbf{q}^{a}\right)}{\partial\mathbf{q}_{\kappa}}\right)
\\
=&\frac{1}{4}\left(\nu\frac{\partial\mathrm{f}\left(\mathbf{q}^{a}\right)}{\partial\mathbf{q}_{r}}+\nu\imath^{\nu^{-1}}\frac{\partial\mathrm{f}\left(\mathbf{q}^{a}\right)}{\partial\mathbf{q}_{\imath}}+\nu\jmath^{\nu^{-1}}\frac{\partial\mathrm{f}}{\partial\mathbf{q}_{\jmath}}+\nu\kappa^{\nu^{-1}}\frac{\partial \mathrm{f}\left(\mathbf{q}^{a}\right)\left(\mathbf{q}^{a}\right)}{\partial\mathbf{q}_{\kappa}}\right)
\\
=&\frac{1}{4}\nu\left(\frac{\partial\mathrm{f}\left(\mathbf{q}^{a}\right)}{\partial\mathbf{q}_{r}}+\imath^{\nu^{-1}}\frac{\partial\mathrm{f}\left(\mathbf{q}^{a}\right)}{\partial\mathbf{q}_{\imath}}+\jmath^{\nu^{-1}}\frac{\partial\mathrm{f}\left(\mathbf{q}^{a}\right)}{\partial\mathbf{q}_{\jmath}}+\kappa^{\nu^{-1}}\frac{\partial \mathrm{f}\left(\mathbf{q}^{a}\right)}{\partial\mathbf{q}_{\kappa}}\right)
\end{aligned}
\label{eq:ImportantReal}
\end{equation}
The expression in \eqref{eq:ImportantReal} is similar to that of \eqref{eq:grad} where the imaginary unit have changed from $\{\imath,\jmath,\kappa\}$ to $\{\imath^{\nu^{-1}},\jmath^{\nu^{1}},\kappa^{\nu^{-1}}\}$. Next we formulate how this change in imaginary units will effect the differentiation base. The differentiation base was $q=q_{r}+\imath q_{\imath}+\jmath q_{\jmath} + \kappa q_{\kappa}$. However, with the changed imaginary units the new differentiation base becomes
\begin{equation}
\begin{aligned}
q_{\text{new}}=&q_{r}+\imath^{\nu^{-1}}q_{\imath}+\jmath^{\nu^{-1}}q_{\jmath}+\kappa^{\nu^{-1}}q_{\kappa}=q_{r}+\nu^{-1}\imath\nu q_{\imath}+\nu^{-1}\jmath\nu q_{\jmath}+\nu^{-1}\kappa\nu q_{\kappa}
\\
=&\nu^{-1}q_{r}\nu+\imath^{\nu^{-1}}q_{\imath}+\jmath^{\nu^{-1}}q_{\jmath}+\kappa^{\nu^{-1}}q_{\kappa}=\nu^{-1}\left(q_{r}+\imath q_{\imath}+\jmath q_{\jmath}+\kappa q_{\kappa}\right)\nu=q^{\nu^{-1}}
\end{aligned}
\label{eq:ImportantRealFianlMapping}
\end{equation}
Therefore, the expression in \eqref{eq:ImportantReal} and \eqref{eq:ImportantRealFianlMapping} indicate
\begin{equation}
\boxed{\begin{aligned}
\frac{\partial\nu\mathrm{f}\left(\mathbf{q}^{a}\right)}{\partial \mathbf{q}^{*}}=&\frac{1}{4}\nu\left(\frac{\partial\mathrm{f}\left(\mathbf{q}^{a}\right)}{\partial\mathbf{q}_{r}}+\imath^{\nu^{-1}}\frac{\partial\mathrm{f}\left(\mathbf{q}^{a}\right)}{\partial\mathbf{q}_{\imath}}+\jmath^{\nu^{-1}}\frac{\partial\mathrm{f}\left(\mathbf{q}^{a}\right)}{\partial\mathbf{q}_{\jmath}}+\kappa^{\nu^{-1}}\frac{\partial \mathrm{f}\left(\mathbf{q}^{a}\right)}{\partial\mathbf{q}_{\kappa}}\right)
\\
=&\nu\frac{\partial\mathrm{f}\left(\mathbf{q}^{a}\right)}{\partial \mathbf{q}^{*}_{\text{new}}}\hspace{0.32cm}\text{with}\hspace{0.32cm}\mathbf{q}_{\text{new}}=\mathbf{q}^{\nu^{-1}}
\end{aligned}}
\label{eq:RealLeftMultiplication}
\end{equation}
Following the same approach, we can now reconstruct the gradient as
\begin{equation}
\nabla_{\mathbf{q}^{a*}}\bigg(\nu\mathrm{f}\left(\mathbf{q}^{a}\right)\bigg)=\nu\bigg(\nabla_{\mathbf{q}^{a
^{\nu^{-1}*}}}\mathrm{f}\left(\mathbf{q}^{a}\right)\bigg)
\end{equation}
Taking into account the non-commutativity of the quaternion product and \eqref{eq:QGO}, for multiplication on the right-hand-side as $\frac{\partial\mathrm{f}\left(\mathbf{q}\right)\nu}{\partial\mathbf{q}^{*}}=\frac{\partial\mathrm{f}\left(\mathbf{q}\right)}{\partial\mathbf{q}^{*}}\nu$, resulting in
\begin{equation}
\nabla_{\mathbf{q}^{a*}} \bigg(\mathrm{f}\left(\mathbf{q}^{a}\right)\nu\bigg)=\bigg(\nabla_{\mathbf{q}^{a*}}(\mathrm{f}\left(\mathbf{q}^{a}\right)\bigg)\nu
\label{eq:RealRightMultiplication}
\end{equation}


\subsection{The Product Rule}

Now we come to the case of multiplications of two quaternion-valued functions. Let us first recall the product rule in $\mathbb{R}$. For real-valued functions $\mathrm{f}\left(\cdot\right):\mathbb{R}\rightarrow\mathbb{R}$ and $\mathrm{p}\left(\cdot\right):\mathbb{R}\rightarrow\mathbb{R}$, the derivate of $\mathrm{h}\left(x\right)=\mathrm{f}\left(x\right)\mathrm{p}\left(x\right)
$ is given by
\begin{equation}
x\in\mathbb{R}:
\frac{\partial\mathrm{f}\left(x\right)\mathrm{p}\left(x\right)}{\partial x}=\underbrace{\frac{\partial\mathrm{f}\left(x\right)}{\partial x}\mathrm{p}\left(x\right)}_{\mathrm{p}\left(x\right)\hspace{0.12cm}\text{treated as a constant}}+\underbrace{\mathrm{f}\left(x\right)\frac{\partial \mathrm{p}\left(x\right)}{\partial x}}_{\mathrm{f}\left(x\right)\hspace{0.12cm}\text{treated as a constant}}
\label{eq:RealMutliRule}
\end{equation}
The expression in \eqref{eq:RealMutliRule} presents the gradient of $\mathrm{h}\left(x\right)=\mathrm{f}\left(x\right)\mathrm{p}\left(x\right)$ as the summation of two components, one treating $\mathrm{f}\left(x\right)$ as a constant, and the other treating $\mathrm{g}\left(x\right)$ as a constant. Let us implement the same concept in the quaternion domain. In this case, for
\begin{equation}
\mathrm{h}\left(\mathbf{q}^{a}\right)=\mathrm{f}\left(\mathbf{q}^{a}\right)\mathrm{p}\left(\mathbf{q}^{a}\right)
\end{equation}
with
\begin{equation}
\mathrm{f}\left(\cdot\right):\mathbb{H}^{m}\rightarrow\mathbb{H}\hspace{0.32cm}\text{and}\hspace{0.32cm}\mathrm{p}\left(\cdot\right):\mathbb{H}^{m}\rightarrow\mathbb{H}
\end{equation}
we have the following two gradient components:
\begin{align}
\text{treating $\mathrm{p}\left(\mathbf{q}^{a}\right)$ as constant:}&\hspace{0.12cm}\frac{\partial \mathrm{h}\left(\mathbf{q}^{a}\right)}{\partial\mathbf{q}^{*}}=\left(\frac{\partial\mathrm{f}\left(\mathbf{q}^{a}\right)}{\partial\mathbf{q}^{*}}\right)\mathrm{p}\left(\mathbf{q}^{a}\right)\label{eq:TwoSections}
\\
\text{treating $\mathrm{f}\left(\mathbf{q}^{a}\right)$ as constant:}&\hspace{0.12cm}\hspace{0.12cm}\frac{\partial \mathrm{h}\left(\mathbf{q}^{a}\right)}{\partial\mathbf{q}^{*}}=\frac{\partial\mathrm{f}\left(\mathbf{q}^{a}\right)\mathrm{p}\left(\mathbf{q}^{a}\right)}{\partial\mathbf{q}^{*}}=\mathrm{f}\left(\mathbf{q}^{a}\right)\frac{\partial\mathrm{p}\left(\mathbf{q}^{a}\right)}{\partial\left(\mathbf{q}^{\left(\mathrm{f}\left(\mathbf{q}^{a}\right)\right)^{-1}}\right)^{*}}\nonumber
\end{align}
Therefore, gradient of $\mathrm{h}\left(\cdot\right):\mathbb{H}^{m}\rightarrow\mathbb{H}$ is given by~\cite{GenHR}
\begin{equation}
\boxed{\frac{\partial\mathrm{h}\left(\mathbf{q}^{a}\right)}{\partial\mathbf{q}^{*}}=\mathrm{f}\left(\mathbf{q}^{a}\right)\frac{\partial\mathrm{p}\left(\mathbf{q}^{a}\right)}{\partial\left(\mathbf{q}^{\left(\mathrm{f}\left(\mathbf{q}^{a}\right)\right)^{-1}}\right)^{*}}+\frac{\partial\mathrm{f}\left(\mathbf{q}^{a}\right)}{\partial\mathbf{q}^{*}}\mathrm{p}\left(\mathbf{q}^{a}\right)}
\label{eq:MultiplicaationRuleFunction}
\end{equation}

\begin{example}
Let us reconsider the derivative of $\mathrm{h}\left(q\right)=\|q\|^{2}$. We can formulate the function as a product, i.e. $\mathrm{h}\left(q\right)=\mathrm{f}\left(q\right)\mathrm{p}\left(q\right)$ with $\mathrm{f}\left(q
\right)=q$ and $\mathrm{p}\left(q\right)=q^{*}$. Using the chain derivative rule we can write
\begin{equation}
\frac{\partial\mathrm{h}\left(q\right)}{\partial q^{*}}=\frac{\partial qq^{*}}{\partial q^{*}}=\frac{\partial q}{\partial q^{*}}q^{*}+q\frac{\partial q^{*}}{\partial\left( q^{q^{-1}}\right)^{*}}
\label{eq:1}
\end{equation}
Let us now focus on the term $\frac{\partial q^{*}}{\partial\left( q^{q^{-1}}\right)^{*}}$. From \eqref{eq:RealLeftMultiplication} it follows that
\begin{equation}
\frac{\partial q^{*}}{\partial\left( q^{q^{-1}}\right)^{*}}=\frac{1}{4}\left(\frac{\partial q^{*}}{\partial q_{r}} +\imath^{q^{-1}}\frac{\partial q^{*}}{\partial q_{\imath}} + \jmath^{q^{-1}}\frac{\partial q^{*}}{\partial q_{\jmath}}+\kappa^{q^{-1}}\frac{\partial q^{*}}{\partial q_{\kappa}}\right)
\label{eq:2}
\end{equation}
In addition we have,
\begin{equation}
\begin{aligned}
\frac{\partial q^{*}}{\partial q_{r}}=&\frac{\partial q_{r}}{\partial q_{r}}-\imath\frac{\partial q_{\imath}}{\partial q_{r}}-\jmath\frac{\partial q_{\jmath}}{\partial q_{r}}-\kappa\frac{\partial q_{\kappa}}{\partial q_{r}}=1
\\
\frac{\partial q^{*}}{\partial q_{\imath}}=&\frac{\partial q_{r}}{\partial q_{\imath}}-\imath\frac{\partial q_{\imath}}{\partial q_{\imath}}-\jmath\frac{\partial q_{\jmath}}{\partial q_{\imath}}-\kappa\frac{\partial q_{\kappa}}{\partial q_{\imath}}=-\imath
\\
\frac{\partial q^{*}}{\partial q_{\jmath}}=&\frac{\partial q_{r}}{\partial q_{\jmath}}-\imath\frac{\partial q_{\imath}}{\partial q_{\jmath}}-\jmath\frac{\partial q_{\jmath}}{\partial q_{\jmath}}-\kappa\frac{\partial q_{\kappa}}{\partial q_{\jmath}}=-\jmath
\\
\frac{\partial q^{*}}{\partial q_{\kappa}}=&\frac{\partial q_{r}}{\partial q_{\kappa}}-\imath\frac{\partial q_{\imath}}{\partial q_{\kappa}}-\jmath\frac{\partial q_{\jmath}}{\partial q_{\kappa}}-\kappa\frac{\partial q_{\kappa}}{\partial q_{r}}=-\kappa
\end{aligned}
\label{eq:3}
\end{equation}
Now, substituting \eqref{eq:3} into \eqref{eq:2} gives
\begin{equation}
\frac{\partial q^{*}}{\partial\left( q^{q^{-1}}\right)^{*}}=\frac{1}{4}\left(1+\imath^{q^{-1}}\left(-\imath\right)+\jmath^{q^{-1}}\left(-\jmath\right)+\kappa^{q^{-1}}\left(-\kappa\right)\right)
\label{eq:4}
\end{equation}
On the other hand, we have
\begin{equation}
\imath^{q^{-1}}\left(-\imath\right)= q^{-1}\imath q\left(-\imath\right)=q^{-1}\left(\imath q \imath^{-1}\right)=q^{-1}\left(q^{\imath}\right)
\label{eq:sub1}
\end{equation}
with the proofs for 
\begin{equation}
\jmath^{q^{-1}}\left(-\jmath\right)=q^{-1}\left(q^{\jmath}\right)\hspace{0.32cm}\text{and}\hspace{0.32cm}\kappa^{q^{-1}}\left(-\kappa\right)=q^{-1}\left(q^{\kappa}\right)
\label{eq:sub2}
\end{equation}
following in the same line of logic. The expressions in \eqref{eq:sub1} and \eqref{eq:sub2} in conjunction with \eqref{eq:4} result in 
\begin{equation}
\begin{aligned}
\frac{\partial q^{*}}{\partial\left( q^{q^{-1}}\right)^{*}}=&\frac{1}{4}\left(1+q^{-1}q^{\imath}+{q^{-1}}q^{\jmath}+q^{-1}q^{\kappa}\right)
\\
=&\frac{1}{4}\left(q^{-1}\right)\left(q+q^{\imath}+q^{\jmath}+q^{\kappa}\right)=q^{-1}q_{r}
\end{aligned}
\label{eq:4-}
\end{equation}
Finally, using \eqref{eq:grad} we can write 
\begin{equation}
\frac{\partial q}{\partial q^{*}}=\frac{1}{4}\left(\frac{\partial q}{\partial q_{r}} +\imath\frac{\partial q}{\partial q_{\imath}} + \jmath\frac{\partial q}{\partial q_{\jmath}}+\kappa\frac{\partial q}{\partial q_{\kappa}}\right)=-\frac{1}{2}
\label{eq:5}
\end{equation}
Replacing \eqref{eq:4-} and \eqref{eq:5} into \eqref{eq:1} gives
\begin{equation}
\frac{\partial\mathrm{h}\left(q\right)}{\partial q^{*}}=\frac{\partial qq^{*}}{\partial q^{*}}=\frac{\partial q}{\partial q^{*}}q^{*}+q\frac{\partial q^{*}}{\partial\left( q^{q^{-1}}\right)^{*}}=-\frac{1}{2}q^{*}+q_{r}=\frac{1}{2}q
\label{eq:finalq2}
\end{equation}
\end{example}



\subsection{Chain Derivative Rule in the $\mathbb{HR}$-Calculus}

Once more, recall the chain derivative rule for real-valued functions. To this end, consider
\begin{equation}
\mathrm{h}\left(x\right)=\mathrm{f}\left(\mathrm{p}\left(x\right)\right)
\end{equation}
with
\begin{equation}
\mathrm{f}\left(\cdot\right):\mathbb{R}\rightarrow\mathbb{R}\hspace{0.12cm}\text{and}\hspace{0.12cm}\mathrm{p}\left(\cdot\right):\mathbb{R}\rightarrow\mathbb{R}
\end{equation}
The goal is to find the rate of change in $\mathrm{h}\left(x\right)=\mathrm{f}\left(\mathrm{p}\left(x\right)\right)$ with respect to changes in $x$, i.e. $\frac{\partial\mathrm{h}\left(x\right)}{\partial x}$. From classical calculus, we have 
\begin{equation}
\frac{\partial\mathrm{h}\left(x\right)}{\partial x}=\frac{\partial\mathrm{p}\left(x\right)}{\partial x}\frac{\partial\mathrm{f}\left(\mathrm{p}\left(x\right)\right)}{\partial\mathrm{p}\left(x
\right)}
\label{ea:Real-Chain}
\end{equation}
consisting of two sections;

\begin{itemize}

\item[i)] the partial derivative
\begin{equation}
\frac{\partial\mathrm{p}\left(x\right)}{\partial x}
\label{eq:chain1}
\end{equation}
represents the rate of change of the inner function $\mathrm{p}\left(\cdot\right)$ with respect to $x$ and captures how much the input to $\mathrm{f}\left(\cdot\right)$ varies as $x$ varies.

\item[ii)] the partial derivative  
\begin{equation}
\frac{\partial\mathrm{f}\left(\mathrm{p}\left(x\right)\right)}{\partial\mathrm{p}\left(x
\right)}
\label{eq:chain2}
\end{equation}
represents the rate of change of the outer function $\mathrm{f}\left(\cdot\right)$ evaluated at $\mathrm{p}\left(\cdot\right)$ and tells us how changes in $\mathrm{p}\left(\cdot\right)$ influence the final output of $\mathrm{h}\left(\cdot\right)$.

\end{itemize}

\noindent Together, these terms describe how a small change in $x$ propagates through the composition $\mathrm{h}\left(\cdot\right)=\mathrm{f}\left(\mathrm{p}\left(\cdot\right)\right)$, first altering the intermediate value $\mathrm{p}\left(\cdot\right)$, and then, affecting the output via passing through $\mathrm{f}\left(\cdot\right)$.

Now, let us come back to quaternion-valued functions. Consider the quaternion-valued compound function $\mathrm{h}\left(\cdot\right)=\mathrm{f}\left(\mathrm{p}\left(\cdot\right)\right):\mathbb{H}^{m}\rightarrow\mathbb{H}$, where
\begin{equation}
\mathrm{p}\left(\cdot\right):\mathbb{H}^{m}\rightarrow\mathbb{H}^{m}\hspace{0.32cm}\text{and}\hspace{0.32cm}\mathrm{f}\left(\cdot\right):\mathbb{H}^{m}\rightarrow\mathbb{H}
\end{equation}
Using the relation in \eqref{eq:QGO} the gradient of $\mathrm{h}\left(\mathbf{q}\right)$ is mapped to the gradient of its real-valued quadrivatiate dual (akin to the process described in \eqref{eq:DefG}), where the classical chain derivative can be used to find the derivatives of $\{\mathrm{h}_{r}\left(\mathbf{q}\right),\mathrm{h}_{\imath}\left(\mathbf{q}\right),\mathrm{h}_{\jmath}\left(\mathbf{q}\right),\mathrm{h}_{\kappa}\left(\mathbf{q}\right)\}$. These derivatives are then mapped back into their quaternion formulation using \eqref{eq:QGO}. Detailed derivations can be found in \cite{GenHR}. However, the important result is





\begin{equation}
\begin{aligned}
\frac{\partial\mathrm{f}\left(\mathrm{p}\left(\mathbf{q}\right)\right)}{\partial\mathbf{q}^{*}}=&\frac{\partial\mathrm{p}^{*}\left(\mathbf{q}\right)}{\partial\mathbf{q}^{*}}\frac{\partial\mathrm{f}\left(\mathrm{p}\left(\mathbf{q}\right)\right)}{\partial\mathrm{
p}^{*}\left(\mathbf{q}\right)}+\frac{\partial\mathrm{p}^{\imath*}\left(\mathbf{q}\right)}{\partial\mathbf{q}^{*}}\frac{\partial\mathrm{f}\left(\mathrm{p}\left(\mathbf{q}\right)\right)}{\partial\mathrm{p}^{\imath*}\left(\mathbf{q}\right)}
\\
&+\frac{\partial\mathrm{p}^{\jmath*}\left(\mathbf{q}\right)}{\partial\mathbf{q}^{*}}\frac{\partial\mathrm{f}\left(\mathrm{p}\left(\mathbf{q}\right)\right)}{\partial\mathrm{p}^{\jmath*}\left(\mathbf{q}\right)}+\frac{\partial\mathrm{p}^{\kappa*}\left(\mathbf{q}\right)}{\partial\mathbf{q}^{*}}\frac{\partial\mathrm{f}\left(\mathrm{p}\left(\mathbf{q}\right)\right)}{\partial\mathrm{p}^{\kappa*}\left(\mathbf{q}\right)}
\end{aligned}
\end{equation}
    

\noindent Perhaps, the most useful simplification in the context of adaptive information processing applications is the special case where $\mathrm{p}\left(\cdot\right):\mathbb{H}^{m}\rightarrow\mathbb{R}$ and $\mathrm{f}\left(\cdot\right):\mathbb{R}\rightarrow\mathbb{R}$. In this scenario, the described process yields

\begin{equation}
\boxed{
\frac{\partial\mathrm{f}\left(\mathrm{p}\left(\mathbf{q}\right)\right)}{\partial\mathbf{q}^{*}}=\frac{\partial\mathrm{p}^{*}\left(\mathbf{q}\right)}{\partial\mathbf{q}^{*}}\frac{\partial\mathrm{f}\left(\mathrm{p}\left(\mathbf{q}\right)\right)}{\partial\mathrm{p}^{*}\left(\mathbf{q}\right)}
\label{eq:RealChain}}
\end{equation}
\\Finally, note that the chain rule derivative is crucial in deriving the quaternion backpropagation algorithm. The complete derivation falls beyond the scope of this chapter; however, the keen reader is referred to \cite{QBA,QBDX}.

\begin{example} Let us now consider the quaternion formulation of the least mean square algorithm, i.e. quaternion least mean square (QLMS) initially proposed by \cite{QLMSCCT2009}. The aim of QLMS is to find weight vector parameters $\mathbf{w}_{\left[n\right]}$ that for the input sequence $\{\mathbf{q}_{[n]}:n\in\mathbb{N}\}$, allows the estimates $\hat{y}_{\left[n\right]}=\mathbf{w}^{\T}_{\left[n\right]}\mathbf{q}^{a}_{\left[n\right]}$ to track observation sequence $\{y_{\left[n\right]}:n\in\mathbb{N}\}$ so that its estimates minimise the the squared error cost given by 
\begin{equation}
J_{\left[n\right]}=\|\epsilon_{\left[n\right]}\|^2
\end{equation}
where $\epsilon_{\left[n\right]}$ is the error at time instant $n$ and is given by \begin{equation}
\epsilon_{\left[n\right]}=y_{\left[n\right]}-\hat{y}_{\left[n\right]}
\label{eq:QLMS-Cost}
\end{equation}
The QLMS achieves the stated goal through iterative updates of the weight vector using the cost function gradient as
\begin{equation}
\mathbf{w}_{\left[n+1\right]}=\mathbf{w}_{\left[n\right]}-\gamma\frac{\partial J_{\left[n\right]}}{\partial \mathbf{w}^{*}_{\left[n\right]}}
\end{equation}
where $\gamma\in\mathbb{R}^{+}$ is an adaptation gain. This requires the steepest direction of change for $J_{\left[n\right]}$ with respect to the weight vector $\mathbf{w}_{\left[n\right]}$, i.e. $\frac{\partial J_{\left[n\right]}}{\partial \mathbf{w}^{*}_{\left[n\right]}}$, to be established. To do this, let us consider the cost function as a composite function so that 
\begin{equation}
J_{\left[n\right]}=\mathrm{f}\left(\mathrm{p}\left(\mathbf{w}_{\left[n\right]}\right)\right)
\end{equation}
with 
\begin{equation}
\begin{aligned}
\mathrm{p}\left(\mathbf{w}_{\left[n\right]}\right)=&\epsilon_{\left[n\right]}=y_{\left[n\right]}-\hat{y}_{\left[n\right]}=y_{\left[n\right]}-\mathbf{w}^{\T}_{\left[n\right]}\mathbf{q}^{a}_{\left[n\right]}
\\
\mathrm{f}(\epsilon_{[n]})=&\epsilon_{[n]}\epsilon^{*}_{[n]}
\end{aligned}
\end{equation}
This fits within the conditions of the chain derivative rule in \eqref{eq:RealChain}. In addition, from \eqref{eq:finalq2}, we know that
\begin{equation}
\frac{\partial\mathrm{f}\left(\epsilon_{[n]}\right)}{\partial\epsilon^{*}_{[n]}}=\frac{\partial\epsilon_{[n]}\epsilon^{*}_{[n]}}{\partial\epsilon^{*}_{[n]}}=\frac{1}{2}\epsilon_{[n]}
\label{eq:first}
\end{equation}
We also consider the involution
\begin{equation}
\frac{1}{2}\epsilon^{\imath}_{\left[n\right]}=\frac{1}{2}\imath\epsilon_{\left[n\right]}\imath^{-1}=\imath\frac{1}{2}\epsilon_{\left[n\right]}\imath^{-1}=\imath\frac{\partial\epsilon_{[n]}\epsilon^{*}_{[n]}}{\partial\epsilon^{*}_{[n]}}\imath^{-1}
\label{eq:11}
\end{equation}
where we have used the relation in \eqref{eq:first}. Now using the derived multiplication rules in \eqref{eq:RealLeftMultiplication} and \eqref{eq:RealRightMultiplication}, the expression in \eqref{eq:11} can be formulated into 
\begin{equation}
\frac{1}{2}\epsilon^{\imath}_{\left[n\right]}=\imath\frac{\partial\epsilon_{[n]}\epsilon^{*}_{[n]}}{\partial\epsilon^{*}_{[n]}}\imath^{-1}=\imath\frac{\partial\epsilon_{[n]}\epsilon^{*}_{[n]}\imath^{-1}}{\partial\epsilon^{*}_{[n]}}=\frac{\partial\imath\epsilon_{[n]}\epsilon^{*}_{[n]}\imath^{-1}}{\partial\epsilon^{\imath*}_{[n]}}
\label{eq:112}
\end{equation}
Considering the fact that $\epsilon_{[n]}\epsilon^{*}_{[n]}\in\mathbb{R}$, it follows that $\imath\epsilon_{[n]}\epsilon^{*}_{[n]}\imath^{-1}=\epsilon_{[n]}\epsilon^{*}_{[n]}$; thus, \eqref{eq:112} gives
\begin{equation}
\frac{1}{2}\epsilon^{\imath}_{\left[n\right]}=\frac{\partial\imath\epsilon_{[n]}\epsilon^{*}_{[n]}\imath^{-1}}{\partial\epsilon^{\imath*}_{[n]}}=\frac{\partial\epsilon_{[n]}\epsilon^{*}_{[n]}}{\partial\epsilon^{\imath*}_{[n]}}
\end{equation}
In a similar fashion it can be shown that 
\begin{equation}
\frac{1}{2}\epsilon^{\jmath}_{\left[n\right]}=\frac{\partial\epsilon_{[n]}\epsilon^{*}_{[n]}}{\partial\epsilon^{\jmath*}_{[n]}}\hspace{0.32cm}\text{and}\hspace{0.32cm}\frac{1}{2}\epsilon^{\kappa}_{\left[n\right]}=\frac{\partial\epsilon_{[n]}\epsilon^{*}_{[n]}}{\partial\epsilon^{\kappa*}_{[n]}}
\label{eq:1fianl1}
\end{equation}

Now for $\xi\in\{1,\imath,\jmath,\kappa\}$, we have to calculate $\frac{\partial\mathrm{p}^{\xi*}\left(\mathbf{w}_{[n]}\right)}{\partial\mathbf{w}^{*}_{[n]}}$. Let us start with a simple case, that is,
\begin{equation}
\begin{aligned}
\frac{\partial\epsilon^{*}_{[n]}}{\partial\mathbf{w}^{*}_{[n]}}=&\frac{\partial y^{*}_{[n]}-\mathbf{q}^{a\H}_{[n]}\mathbf{w}^{*}_{[n]}}{\partial\mathbf{w}^{*}_{\left[n\right]}}=-\frac{\partial\mathbf{q}^{a\H}_{[n]}\mathbf{w}^{*}
_{[n]}}{\partial\mathbf{w}^{*}_{[n]}}
\\
=&-\frac{1}{4}\left(\frac{\partial\mathbf{q}^{a\H}_{[n]}\mathbf{w}^{*}_{[n]}}{\partial\mathbf{w}_{r_{[n]}}}+\imath\frac{\partial\mathbf{q}^{a\H}_{[n]}\mathbf{w}^{*}_{[n]}}{\partial\mathbf{w}_{\imath_{[n]}}}+\jmath\frac{\partial\mathbf{q}^{a\H}_{[n]}\mathbf{w}^{*}_{[n]}}{\partial\mathbf{w}_{\jmath_{[n]}}}+\kappa\frac{\partial\mathbf{q}^{a\H}_{[n]}\mathbf{w}^{*}_{[n]}}{\partial\mathbf{w}_{\kappa_{[n]}}}\right)
\\
=&-\frac{1}{4}\left(\mathbf{q}^{a*}+\imath\mathbf{q}^{a*}(-\imath)+\jmath\mathbf{q}^{a*}(-\jmath)+\kappa\mathbf{q}^{a*}(-\kappa)\right)=-\begin{bmatrix}\mathbf{q}_{r}\\\mathbf{q}_{r}\\\mathbf{q}_{r}\\\mathbf{q}_{r}\end{bmatrix}
\end{aligned}
\end{equation}
In addition, for the case of $\frac{\partial\mathrm{p}^{\imath*}\left(\mathbf{w}_{[n]}\right)}{\partial\mathbf{w}^{\imath*}}$, we can write
\begin{equation}
\begin{aligned}
\frac{\partial\epsilon^{\imath*}_{[n]}}{\partial\mathbf{w}^{*}_{[n]}}=&\frac{\partial y^{\imath*}_{[n]}-\mathbf{q}^{a\imath\H}_{[n]}\mathbf{w}^{\imath*}_{[n]}}{\partial\mathbf{w}^{*}_{\left[n\right]}}=-\frac{\partial\mathbf{q}^{a\imath\H}_{[n]}\mathbf{w}^{\imath*}
_{[n]}}{\partial\mathbf{w}^{\imath*}_{[n]}}
\\
=&-\frac{1}{4}\left(\frac{\partial\mathbf{q}^{a\imath\H}_{[n]}\mathbf{w}^{\imath*}_{[n]}}{\partial\mathbf{w}_{r_{[n]}}}+\imath\frac{\partial\mathbf{q}^{a\imath\H}_{[n]}\mathbf{w}^{\imath*}_{[n]}}{\partial\mathbf{w}_{\imath_{[n]}}}+\jmath\frac{\partial\mathbf{q}^{a\imath\H}_{[n]}\mathbf{w}^{\imath*}_{[n]}}{\partial\mathbf{w}_{\jmath_{[n]}}}+\kappa\frac{\partial\mathbf{q}^{a\imath\H}_{[n]}\mathbf{w}^{\imath*}_{[n]}}{\partial\mathbf{w}_{\kappa_{[n]}}}\right)
\\
=&-\frac{1}{4}\left(\mathbf{q}^{a\imath*}+\mathbf{q}^{a*}-\mathbf{q}^{a\kappa*}-\mathbf{q}^{a\jmath*}\right)=\begin{bmatrix}{\color{white}-}\mathbf{q}_{\imath}\imath\\{\color{white}-}\mathbf{q}_{\imath}\imath\\-\mathbf{q}_{\imath}\imath\\-\mathbf{q}_{\imath}\imath\end{bmatrix}
\end{aligned}
\end{equation}
In a similar fashion we can show that
\begin{equation}
\frac{\partial\epsilon^{\jmath*}_{[n]}}{\partial\mathbf{w}^{*}_{[n]}}=\begin{bmatrix}{\color{white}-}\mathbf{q}_{\jmath}\jmath\\-\mathbf{q}_{\jmath}\jmath\\{\color{white}-}\mathbf{q}_{\jmath}\jmath\\-\mathbf{q}_{\imath}\jmath\end{bmatrix}\hspace{0.32cm}\text{and}\hspace{0.32cm}\frac{\partial\epsilon^{\kappa*}_{[n]}}{\partial\mathbf{w}^{*}_{[n]}}=\begin{bmatrix}{\color{white}-}\mathbf{q}_{\kappa}\kappa\\-\mathbf{q}_{\kappa}\kappa\\-\mathbf{q}_{\kappa}\kappa\\{\color{white}-}\mathbf{q}_{\kappa}\kappa\end{bmatrix}
\label{eq:last}
\end{equation}
Substituting \eqref{eq:first}-\eqref{eq:last} into the chain derivative gives the overall differential as
\begin{equation}
\frac{\partial J_{[n]}}{\partial\mathbf{w}^{*}_{[n]}}=-\frac{1}{2}\epsilon_{[n]}\mathbf{q}^{a*}_{[n]}
\end{equation}
Thus, the weight update term becomes
\begin{equation}
\mathbf{w}_{\left[n+1\right]}=\mathbf{w}_{\left[n\right]}+\gamma\epsilon_{\left[n\right]}\mathbf{q}^{a*}_{\left[n\right]}
\label{eq:GradLeft}
\end{equation}
with the $\frac{1}{2}$ factor absorbed into the adaptation gain.  Alternatively, using the product rule in \eqref{eq:MultiplicaationRuleFunction}, we can also have
\begin{equation}
\begin{aligned}
\frac{\partial J_{\left[n\right]}}{\partial\mathbf{w}^{*}_{\left[n\right]}}=&\frac{\partial\|\epsilon_{\left[n\right]}\|^{2}}{\partial\mathbf{w}^{*}_{\left[n\right]}}=\frac{\partial\left(\epsilon^{*}_{\left[n\right]}\epsilon_{\left[n\right]}\right)}{\partial\mathbf{w}^{*}_{\left[n\right]}}=\epsilon^{*}_{\left[n\right]}\frac{\partial\epsilon_{\left[n\right]}}{\partial\left(\mathbf{w}^{\epsilon^{*^{-1}}_{[n]}}_{[n]}\right)^{*}}+\frac{\partial\epsilon^{*}_{\left[n\right]}}{\partial\mathbf{w}^{*}_{\left[n\right]}}\epsilon_{\left[n\right]}
\\
=&\frac{1}{2}\epsilon_{\left[n\right]}\mathbf{q}^{a}_{\left[n\right]}-\begin{bmatrix}\mathbf{q}_{r}\\\mathbf{q}_{r}\\\mathbf{q}_{r}\\\mathbf{q}_{r}\end{bmatrix}\epsilon_{\left[n\right]}=-\frac{1}{2}\epsilon_{\left[n\right]}\mathbf{q}^{a*}_{\left[n\right]}
\end{aligned}
\label{eq:GradRight}
\end{equation}
The MATLAB code\footnote{The involution function invijk(q,`i') in line 14 implements the $\imath-$involution, which can be downloaded at \href{https://www.researchgate.net/publication/327765279_Quaternion_Statistics}{here}.} for the QLMS is given below.
\begin{lstlisting}[
frame=single,
numbers=left,
style=Matlab-editor, basicstyle=\footnotesize]
% Quaternion Least Mean Square Algorithm
% The algorithm requires values for: 
% i) mu (the adaptation gian),
% ii) q (an input sequence),
% iii) y (desired or refrence signal).

% The number of iterations
N = length(q); 
% Initial weight vector comprising of 4 weights
w=quaternion(zeros(4,1),zeros(4,1),zeros(4,1),zeros(4,1));

for n = 1:N
   % Formulating the augmented vector in Eq. (1.31)
    qa = [q(n);             invijk(q(n),'i') ;...
          invijk(q(n),'j') ;invijk(q(n),'k') ];
   % Calculating the estimate of the reference signal
   yest = transpose(w)*qa; 
   % Calculating the error signal
   e = y(n) - yest; 
   % Updating the weight vector   
   w = w + mu*e*conj(qa); 
end

\end{lstlisting}
\end{example}
\begin{example} We can now consider the nonlinear quaternion-valued filtering and learning problem. To this end, the aim is to find the weight vector $\mathbf{w}_{\left[n\right]}$ that allows the estimates $\hat{y}_{\left[n\right]}=\phi\left(\mathbf{w}^{\T}_{\left[n\right]}\mathbf{q}^{a}_{\left[n\right]}\right)$ to track observation sequence $\{y_{\left[n\right]}:n\in\mathbb{N}\}$ so that its estimates minimise the squared error cost given by 
\begin{equation}
J_{\left[n\right]}=\|\epsilon_{\left[n\right]}\|^2
\end{equation}
where $\epsilon_{\left[n\right]}$ is the error at time instant $n$ and is given by \begin{equation}
\epsilon_{\left[n\right]}=y_{\left[n\right]}-\hat{y}_{\left[n\right]}
\label{eq:QLMS-Cost}
\end{equation}
with $\phi\left(\cdot\right):\mathbb{H}\rightarrow\mathbb{H}$ denoting a general nonlinear quaternion-valued function. 

In a similar fashion to the QLMS, the quadratic error minimization is achieved through the iterative updates of the weight vector as
\begin{equation}
\mathbf{w}_{\left[n+1\right]}=\mathbf{w}_{\left[n\right]}-\gamma\frac{\partial J_{\left[n\right]}}{\partial \mathbf{w}^{*}_{\left[n\right]}}
\end{equation}
where $\gamma\in\mathbb{R}^{+}$ is an adaptation gain. Once more, we need to evaluate the steepest direction of change for $J_{\left[n\right]}$ with respect to the weight vector $\mathbf{w}_{\left[n\right]}$, i.e. $\frac{\partial J_{\left[n\right]}}{\partial \mathbf{w}^{*}_{\left[n\right]}}$, to be established. To do this, using the chain derivative rule, we have
%
%
\begin{equation}
\frac{\partial J_{\left[n\right]}}{\partial\mathbf{w}^{*}_{\left[n\right]}}= 
\frac{\partial\left(\epsilon^{*}_{\left[n\right]}\epsilon_{\left[n\right]}\right)}{\partial\mathbf{w}^{*}_{\left[n\right]}}=\epsilon^{*}_{\left[n\right]}\frac{\partial\epsilon_{\left[n\right]}}{\partial\left(\mathbf{w}^{\epsilon^{*^{-1}}_{[n]}}_{[n]}\right)^{*}}+\frac{\partial\epsilon^{*}_{\left[n\right]}}{\partial\mathbf{w}^{*}_{\left[n\right]}}\epsilon_{\left[n\right]}
\label{eq:nonlinear1}
\end{equation}
\noindent To evaluate the expression in \eqref{eq:nonlinear1}, we have to evaluate
\begin{equation}
\frac{\partial\epsilon^{*}_{\left[n\right]}}{\partial\mathbf{w}^{*}_{\left[n\right]}}\hspace{0.32cm}\text{and}\hspace{0.32cm}\frac{\partial\epsilon_{\left[n\right]}}{\partial\left(\mathbf{w}^{\epsilon^{*^{-1}}_{[n]}}_{[n]}\right)^{*}}
\end{equation}
Let us start with $\frac{\partial\epsilon^{*}_{\left[n\right]}}{\partial\mathbf{w}^{*}_{\left[n\right]}}$. To this end, from the definition in \eqref{eq:grad}, we have
\begin{equation}
\frac{\partial\epsilon^{*}_{\left[n\right]}}{\partial\mathbf{w}^{*}_{\left[n\right]}}=-\frac{1}{4}\left(\begin{aligned}&\frac{\partial\phi^{*}\left(\mathbf{w}^{\T}_{\left[n\right]}\mathbf{q}^{a}_{\left[n\right]}\right)}{\partial\mathbf{w}_{r_{\left[n\right]}}}+\imath\frac{\partial\phi^{*}\left(\mathbf{w}^{\T}_{\left[n\right]}\mathbf{q}^{a}_{\left[n\right]}\right)}{\partial\mathbf{w}_{\imath_{\left[n\right]}}}
\\
&+\jmath\frac{\partial\phi^{*}\left(\mathbf{w}^{\T}_{\left[n\right]}\mathbf{q}^{a}_{\left[n\right]}\right)}{\partial\mathbf{w}_{\jmath_{\left[n\right]}}}+\kappa\frac{\partial\phi^{*}\left(\mathbf{w}^{\T}_{\left[n\right]}\mathbf{q}^{a}_{\left[n\right]}\right)}{\partial\mathbf{w}_{\kappa_{\left[n\right]}}}\end{aligned}\right)
\label{eq:nonlinear2}
\end{equation}
Since, the derivatives on the right-hand-side of \eqref{eq:nonlinear2} are with respect to real-valued variables $\{\mathbf{w}_{r_{\left[n\right]}},\mathbf{w}_{\imath_{\left[n\right]}},\mathbf{w}_{\jmath_{\left[n\right]}},\mathbf{w}_{\kappa_{\left[n\right]}}\}$, traditional derivative rules, that treat the imaginary units as regular variables, apply here. Similarly, using \eqref{eq:RealLeftMultiplication}, we have 
\begin{equation}
\frac{\partial\epsilon_{\left[n\right]}}{\partial\left(\mathbf{w}^{\epsilon^{*^{-1}}_{[n]}}_{[n]}\right)^{*}}=-\frac{1}{4}\left(\begin{aligned}&\frac{\partial\phi\left(\mathbf{w}^{\T}_{\left[n\right]}\mathbf{q}^{a}_{\left[n\right]}\right)}{\partial\mathbf{w}_{r_{\left[n\right]}}}+\imath^{\epsilon^{*^{-1}}_{\left[n\right]}}\frac{\partial\phi\left(\mathbf{w}^{\T}_{\left[n\right]}\mathbf{q}^{a}_{\left[n\right]}\right)}{\partial\mathbf{w}_{\imath_{\left[n\right]}}}
\\
&+\jmath^{\epsilon^{*^{-1}}_{\left[n\right]}}\frac{\partial\phi\left(\mathbf{w}^{\T}_{\left[n\right]}\mathbf{q}^{a}_{\left[n\right]}\right)}{\partial\mathbf{w}_{\jmath_{\left[n\right]}}}+\kappa^{\epsilon^{*^{-1}}_{\left[n\right]}}\frac{\partial\phi\left(\mathbf{w}^{\T}_{\left[n\right]}\mathbf{q}^{a}_{\left[n\right]}\right)}{\partial\mathbf{w}_{\kappa_{\left[n\right]}}}\end{aligned}\right)
\end{equation}
which can be simplified into
\begin{equation}
\frac{\partial\epsilon_{\left[n\right]}}{\partial\left(\mathbf{w}^{\epsilon^{*^{-1}}_{[n]}}_{[n]}\right)^{*}}=-\frac{\epsilon^{*^{-1}}_{\left[n\right]}}{4}\left(\begin{aligned}&\epsilon^{*}_{\left[n\right]}\frac{\partial\phi\left(\mathbf{w}^{\T}_{\left[n\right]}\mathbf{q}^{a}_{\left[n\right]}\right)}{\partial\mathbf{w}_{r_{\left[n\right]}}}+\imath\epsilon^{*}_{\left[n\right]}\frac{\partial\phi\left(\mathbf{w}^{\T}_{\left[n\right]}\mathbf{q}^{a}_{\left[n\right]}\right)}{\partial\mathbf{w}_{\imath_{\left[n\right]}}}
\\
&+\jmath\epsilon^{*}_{\left[n\right]}\frac{\partial\phi\left(\mathbf{w}^{\T}_{\left[n\right]}\mathbf{q}^{a}_{\left[n\right]}\right)}{\partial\mathbf{w}_{\jmath_{\left[n\right]}}}+\kappa\epsilon^{*}_{\left[n\right]}\frac{\partial\phi\left(\mathbf{w}^{\T}_{\left[n\right]}\mathbf{q}^{a}_{\left[n\right]}\right)}{\partial\mathbf{w}_{\kappa_{\left[n\right]}}}\end{aligned}\right)
\label{eq:nonlinear3}
\end{equation}
Once more, given the fact that the derivatives on the right-hand-side of \eqref{eq:nonlinear3} are with respect to real-valued variables $\{\mathbf{w}_{r_{\left[n\right]}},\mathbf{w}_{\imath_{\left[n\right]}},\mathbf{w}_{\jmath_{\left[n\right]}},\mathbf{w}_{\kappa_{\left[n\right]}}\}$, traditional derivative rules, treat the imaginary units as regular variables, apply here. Finally, substituting \eqref{eq:nonlinear2} and \eqref{eq:nonlinear3} into \eqref{eq:nonlinear1} gives 
\begin{equation}
\frac{\partial J_{\left[n\right]}}{\partial\mathbf{w}^{*}_{\left[n\right]}}=\frac{-1}{4}\left(\begin{aligned}&\frac{\partial\phi^{*}\left(\mathbf{w}^{\T}_{\left[n\right]}\mathbf{q}^{a}_{\left[n\right]}\right)}{\partial\mathbf{w}_{r_{\left[n\right]}}}\epsilon_{\left[n\right]}+\epsilon^{*}_{\left[n\right]}\frac{\partial\phi\left(\mathbf{w}^{\T}_{\left[n\right]}\mathbf{q}^{a}_{\left[n\right]}\right)}{\partial\mathbf{w}_{r_{\left[n\right]}}}
\\
&+\imath\left(\frac{\partial\phi^{*}\left(\mathbf{w}^{\T}_{\left[n\right]}\mathbf{q}^{a}_{\left[n\right]}\right)}{\partial\mathbf{w}_{\imath_{\left[n\right]}}}\epsilon_{\left[n\right]}+\epsilon^{*}_{\left[n\right]}\frac{\partial\phi\left(\mathbf{w}^{\T}_{\left[n\right]}\mathbf{q}^{a}_{\left[n\right]}\right)}{\partial\mathbf{w}_{\imath_{\left[n\right]}}}\right)
\\
&+\jmath\left(\frac{\partial\phi^{*}\left(\mathbf{w}^{\T}_{\left[n\right]}\mathbf{q}^{a}_{\left[n\right]}\right)}{\partial\mathbf{w}_{\jmath_{\left[n\right]}}}\epsilon_{\left[n\right]}+\epsilon^{*}_{\left[n\right]}\frac{\partial\phi\left(\mathbf{w}^{\T}_{\left[n\right]}\mathbf{q}^{a}_{\left[n\right]}\right)}{\partial\mathbf{w}_{\jmath_{\left[n\right]}}}\right)
\\
&+\kappa\left(\frac{\partial\phi^{*}\left(\mathbf{w}^{\T}_{\left[n\right]}\mathbf{q}^{a}_{\left[n\right]}\right)}{\partial\mathbf{w}_{\kappa_{\left[n\right]}}}\epsilon_{\left[n\right]}+\epsilon^{*}_{\left[n\right]}\frac{\partial\phi\left(\mathbf{w}^{\T}_{\left[n\right]}\mathbf{q}^{a}_{\left[n\right]}\right)}{\partial\mathbf{w}_{\kappa_{\left[n\right]}}}\right)
\end{aligned}\right)
\label{eq:nonlinear-final-1}
\end{equation}
The expression in \eqref{eq:nonlinear-final-1} can be significantly simplified if the nonlinear function, $\phi\left(\cdot\right)$, is selected to be locally analytic as in (\ref{eq:PolarPresentation}). In this case, we have 
%
\begin{align} 
\frac{\partial J_{\left[n\right]}}{\partial\mathbf{w}^{*}_{\left[n\right]}} &=
-\frac{1}{2}\epsilon_{\left[n\right]}\frac{\partial\phi^*\left(\mathbf{w}^{\T}_{\left[n\right]}\mathbf{q}^{a}_{\left[n\right]}\right)}{\partial\left(\mathbf{w}^{\T}_{\left[n\right]}\mathbf{q}^{a}_{\left[n\right]}\right)}\mathbf{q}^{a*}_{[n]}\\ 
%
  %
    &= -\frac{1}{2}\epsilon_{\left[n\right]}\phi'\left(\mathbf{q}^{a \H}_{\left[n\right]}  
    \mathbf{w}^*_{\left[n\right]}\right)\mathbf{q}^{a*}_{[n]}
 %
\end{align}
where $\phi'(x)$ denotes the derivative of $\phi(x)$ with respect to $x$. The trick of $\phi^*(x)=\phi(x^*)$ is possible due to the local analyticity of the quaternion function; this trick can be useful in more complex derivations of other algorithms. Examples of functions that meet these requirements include $\text{tanh}\left(\cdot\right)$, $\text{tan}\left(\cdot\right)$, $\text{arctan}\left(\cdot\right)$, $\text{sinh}\left(\cdot\right)$, and $\text{sin}\left(\cdot\right)$, with a more detailed analysis of these functions provided in~\cite{QNAF}. The following MATLAB code\footnote{The involution function invijk(q,`i') in line 14 implements the $\imath-$involution, which can be downloaded at \href{https://www.researchgate.net/publication/327765279_Quaternion_Statistics}{here}.} considers the hyperbolic tangent function as the nonlinear function on the output of the QLMS algorithm. 
\begin{lstlisting}[
frame=single,
numbers=left,
style=Matlab-editor, basicstyle=\footnotesize]
% Nonlinear Quaternion Least Mean Square Algorithm
% The algorithm requires values for:
% i) mu (the adaptation gain),
% ii) q (an input sequence),
% iii) y (desired or reference signal).

% Number of iterations
N = length(q); 
% Initial weight vector comprising of 4 weights
w=quaternion(zeros(4,1),zeros(4,1),zeros(4,1),zeros(4,1));

for n = 1:N
   % Formulating the augmented vector in Eq. (1.31)
    qa = [q(n);             invijk(q(n),'i') ;...
          invijk(q(n),'j') ;invijk(q(n),'k') ];
   % Calculating the nonlinear output
    yest = tanh(transpose(w)*qa); 
     e = y(n) - yest; 
   % Updating the weight vector
   w = w + mu*e*( sech( qa'*conj(w) )^2 )*conj(qa);  
end

\end{lstlisting}
\end{example}

\section{Summary}
This chapter presents the advanced theory of quaternions in terms of algebra, statistics, and calculus. Beginning with the history of quaternions, we then introduced the quaternion algebra and showed how quaternions can be used for three-dimensional rotations. In particular, we revisited quaternion involutions and illustrated how they rotate a quaternion random variable. These quaternion involutions were then shown to construct a vector basis, the so-called the `augmented' quaternion approach. This approach enabled a complete description of a random quaternion process. This was shown in the context of second order statistics (i.e. autocorrelations). The same principle can be extended to other statistical descriptors. Next, we presented the hypercomplex widely linear model  for least squares algorithms followed by the $\mathbb{HR}$-calculus for quaternion derivatives. The most common derivative rules, such as the product rule and the chain rule, were revisited. To shed light on these relatively new derivative rules, the quaternion least mean square was derived twice. Throughout this chapter, many illustrative examples as well as MATLAB codes were provided for ease of exposition and reproducibility of the materials presented herein. 
\Backmatter
%
%
%
%
%
%
%
%
%
\section*{References}
\smaller
 \bibliographystyle{elsarticle-harv}
 \bibliography{cas-refs}

\begin{thebibliography}{49}
\expandafter\ifx\csname natexlab\endcsname\relax\def\natexlab#1{#1}\fi
\providecommand{\url}[1]{\texttt{#1}}
\providecommand{\href}[2]{#2}
\providecommand{\path}[1]{#1}
\providecommand{\DOIprefix}{doi:}
\providecommand{\ArXivprefix}{arXiv:}
\providecommand{\URLprefix}{URL: }
\providecommand{\Pubmedprefix}{pmid:}
\providecommand{\doi}[1]{\href{http://dx.doi.org/#1}{\path{#1}}}
\providecommand{\Pubmed}[1]{\href{pmid:#1}{\path{#1}}}
\providecommand{\bibinfo}[2]{#2}
\ifx\xfnm\relax \def\xfnm[#1]{\unskip,\space#1}\fi
\bibitem[{Allenby(1991)}]{AbstractAlgebra}
\bibinfo{author}{Allenby, R.B.}, \bibinfo{year}{1991}.
\newblock \bibinfo{title}{Rings, Fields, and Groups: {An} Introduction to Abstract Algebra}.
\newblock \bibinfo{edition}{2} ed., \bibinfo{publisher}{Oxford University Press}, \bibinfo{address}{Oxford, England}.
\bibitem[{Bourigault et~al.(2024)Bourigault, Xu and Mandic}]{QBDX}
\bibinfo{author}{Bourigault, P.}, \bibinfo{author}{Xu, D.}, \bibinfo{author}{Mandic, D.P.}, \bibinfo{year}{2024}.
\newblock \bibinfo{title}{Quaternion recurrent neural network with real-time recurrent learning and maximum correntropy criterion}, in: \bibinfo{booktitle}{2024 International Joint Conference on Neural Networks (IJCNN)}, pp. \bibinfo{pages}{1--8}.
\newblock \DOIprefix\doi{10.1109/IJCNN60899.2024.10650324}.
\bibitem[{Brasil et~al.(2018)Brasil, de~Leles Ferreira~Filho and Ishihara}]{3B}
\bibinfo{author}{Brasil, V.d.P.}, \bibinfo{author}{de~Leles Ferreira~Filho, A.}, \bibinfo{author}{Ishihara, J.a.Y.}, \bibinfo{year}{2018}.
\newblock \bibinfo{title}{Electrical three phase circuit analysis using quaternions}.
\newblock \bibinfo{journal}{International Conference on Harmonics and Quality of Power} , \bibinfo{pages}{1--6}.
\bibitem[{Brechet(2022)}]{CalssicalRotation}
\bibinfo{author}{Brechet, S.D.}, \bibinfo{year}{2022}.
\newblock \bibinfo{title}{Rotations in classical mechanics using geometric algebra}.
\newblock \bibinfo{journal}{arXiv:2210.16803} .
\bibitem[{Cheong~Took and Mandic(2009)}]{QLMSCCT2009}
\bibinfo{author}{Cheong~Took, C.}, \bibinfo{author}{Mandic, D.P.}, \bibinfo{year}{2009}.
\newblock \bibinfo{title}{The quaternion {LMS} algorithm for adaptive filtering of hypercomplex processes}.
\newblock \bibinfo{journal}{IEEE Transactions on Signal Processing} \bibinfo{volume}{57}, \bibinfo{pages}{1316--1327}.
\bibitem[{{Cheong Took} et~al.(2011){Cheong Took}, Mandic and Zhang}]{CHEONGTOOKijkHermitian}
\bibinfo{author}{{Cheong Took}, C.}, \bibinfo{author}{Mandic, D.P.}, \bibinfo{author}{Zhang, F.}, \bibinfo{year}{2011}.
\newblock \bibinfo{title}{On the unitary diagonalisation of a special class of quaternion matrices}.
\newblock \bibinfo{journal}{Applied Mathematics Letters} \bibinfo{volume}{24}, \bibinfo{pages}{1806--1809}.
\bibitem[{Cheong~Took et~al.(2024)Cheong~Took, Talebi, Fernandez~Alcala and Mandic}]{CliveSPM2024}
\bibinfo{author}{Cheong~Took, C.}, \bibinfo{author}{Talebi, S.P.}, \bibinfo{author}{Fernandez~Alcala, R.M.}, \bibinfo{author}{Mandic, D.P.}, \bibinfo{year}{2024}.
\newblock \bibinfo{title}{Augmented statistics of quaternion random variables: A lynchpin of quaternion learning machines}.
\newblock \bibinfo{journal}{IEEE Signal Processing Magazine} \bibinfo{volume}{41}, \bibinfo{pages}{72--87}.
\newblock \DOIprefix\doi{10.1109/MSP.2024.3384178}.
\bibitem[{Comminiello et~al.(2019)Comminiello, Lella, Scardapane and Uncini}]{3DS}
\bibinfo{author}{Comminiello, D.}, \bibinfo{author}{Lella, M.}, \bibinfo{author}{Scardapane, S.}, \bibinfo{author}{Uncini, A.}, \bibinfo{year}{2019}.
\newblock \bibinfo{title}{Quaternion convolutional neural networks for detection and localization of {3D} sound events}.
\newblock \bibinfo{journal}{In Proceedings of IEEE International Conference on Acoustics, Speech and Signal Processing} , \bibinfo{pages}{8533--8537}.
\bibitem[{Crassidis et~al.(2007)Crassidis, Markley and Cheng}]{Survey}
\bibinfo{author}{Crassidis, J.L.}, \bibinfo{author}{Markley, F.L.}, \bibinfo{author}{Cheng, Y.}, \bibinfo{year}{2007}.
\newblock \bibinfo{title}{Survey of nonlinear attitude estimation methods}.
\newblock \bibinfo{journal}{Journal of Guidance, Control, and Dynamics} \bibinfo{volume}{30}, \bibinfo{pages}{12--28}.
\bibitem[{Desoer and Kuh(2009)}]{Madar-1-2}
\bibinfo{author}{Desoer, C.A.}, \bibinfo{author}{Kuh, E.H.}, \bibinfo{year}{2009}.
\newblock \bibinfo{title}{Basic circuit theory}. volume~\bibinfo{volume}{2}.
\newblock \bibinfo{publisher}{McGraw-Hill}.
\bibitem[{Ell and Sangwine(2007)}]{Q-Invo}
\bibinfo{author}{Ell, T.A.}, \bibinfo{author}{Sangwine, S.J.}, \bibinfo{year}{2007}.
\newblock \bibinfo{title}{Quaternion involutions and anti-involutions}.
\newblock \bibinfo{journal}{Computers \& Mathematics with Applications} \bibinfo{volume}{53}, \bibinfo{pages}{137--143}.
\bibitem[{Finkelstein et~al.(2004)Finkelstein, Jauch, Schiminovich and Speiser}]{QQM}
\bibinfo{author}{Finkelstein, D.}, \bibinfo{author}{Jauch, J.M.}, \bibinfo{author}{Schiminovich, S.}, \bibinfo{author}{Speiser, D.}, \bibinfo{year}{2004}.
\newblock \bibinfo{title}{Foundations of quaternion quantum mechanics}.
\newblock \bibinfo{journal}{Journal of Mathematical Physics} \bibinfo{volume}{3}, \bibinfo{pages}{207--220}.
\newblock \URLprefix \url{https://doi.org/10.1063/1.1703794}, \DOIprefix\doi{10.1063/1.1703794}.
\bibitem[{Fueter(1939)}]{FueterGerman}
\bibinfo{author}{Fueter, R.}, \bibinfo{year}{1939}.
\newblock \bibinfo{title}{\"{U}ber einen hartogs'schen satz.}
\newblock \bibinfo{journal}{Commentarii Mathematici Helvetici} \bibinfo{volume}{12}, \bibinfo{pages}{75--80}.
\bibitem[{Hamilton(1844)}]{HamiltonPaper}
\bibinfo{author}{Hamilton, W.R.}, \bibinfo{year}{1844}.
\newblock \bibinfo{title}{On quaternions, or on a new system of imaginaries in algebra} \bibinfo{volume}{25}, \bibinfo{pages}{10–13}.
\bibitem[{Hampshire(2018)}]{HeavisideIntro}
\bibinfo{author}{Hampshire, D.P.}, \bibinfo{year}{2018}.
\newblock \bibinfo{title}{A derivation of {M}axwell's equations using the {H}eaviside notation}.
\newblock \bibinfo{journal}{Philosophical Transactions of the Royal Society A: Mathematical, Physical and Engineering Sciences} .
\bibitem[{Hankins(1977)}]{Letter}
\bibinfo{author}{Hankins, T.L.}, \bibinfo{year}{1977}.
\newblock \bibinfo{title}{Triplets and triads: {Sir William Rowan Hamilton} on the metaphysics of mathematics}.
\newblock \bibinfo{journal}{Isis} \bibinfo{volume}{68}, \bibinfo{pages}{175--193}.
\newblock \URLprefix \url{http://www.jstor.org/stable/230069}.
\bibitem[{Heaviside(1920)}]{Heaviside}
\bibinfo{author}{Heaviside, O.}, \bibinfo{year}{1920}.
\newblock \bibinfo{title}{Electromagnetic Theory}. volume~\bibinfo{volume}{2}.
\newblock \bibinfo{publisher}{The Electrician Printing and Publishing Company}.
\newblock \bibinfo{note}{{Reprinted by Dover in 1950}}.
\bibitem[{Kliuchnikov and Yard(2015)}]{QuantComplier}
\bibinfo{author}{Kliuchnikov, V.}, \bibinfo{author}{Yard, J.T.}, \bibinfo{year}{2015}.
\newblock \bibinfo{title}{A framework for exact synthesis}.
\newblock \bibinfo{type}{Technical Report}.
\newblock \bibinfo{note}{ArXiv preprint arXiv:1504.04350}.
\bibitem[{{Kreutz–Delgado}(2006)}]{Ken}
\bibinfo{author}{{Kreutz–Delgado}, K.}, \bibinfo{year}{2006}.
\newblock \bibinfo{title}{The complex gradient operator and the $\mathbb{CR}$-calculus}.
\newblock \bibinfo{publisher}{University of California, San Diego, Technical Report}.
\bibitem[{Kuipers(1999)}]{Kuipers}
\bibinfo{author}{Kuipers, J.B.}, \bibinfo{year}{1999}.
\newblock \bibinfo{title}{Quaternions and Rotation Sequences: {A} Primer with Applications to Orbits, Aerospace and Virtual Reality}.
\newblock \bibinfo{publisher}{Princeton University Press}.
\bibitem[{Leo and Rotelli(2003)}]{Leo}
\bibinfo{author}{Leo, S.D.}, \bibinfo{author}{Rotelli, P.}, \bibinfo{year}{2003}.
\newblock \bibinfo{title}{Quaternion analyticity}.
\newblock \bibinfo{journal}{Applied Mathematics Letters} \bibinfo{volume}{16}, \bibinfo{pages}{1077–1081}.
\bibitem[{Mandic and Goh(2009)}]{DB}
\bibinfo{author}{Mandic, D.P.}, \bibinfo{author}{Goh, V.S.L.}, \bibinfo{year}{2009}.
\newblock \bibinfo{title}{Complex valued nonlinear adaptive filters: Noncircularity, widely linear and neural models}.
\newblock \bibinfo{publisher}{Wielly}.
\bibitem[{Maxwell(1873)}]{MaxEqu}
\bibinfo{author}{Maxwell, J.C.}, \bibinfo{year}{1873}.
\newblock \bibinfo{title}{A Treatise on Electricity and Magnetism}.
\newblock \bibinfo{publisher}{Dover Publications}, \bibinfo{address}{New York}.
\newblock \bibinfo{note}{ISBN 0-486-60636-8 (Vol. 1) and 0-486-60637-6 (Vol. 2)}.
\bibitem[{Nos(2016)}]{Oleg}
\bibinfo{author}{Nos, O.V.}, \bibinfo{year}{2016}.
\newblock \bibinfo{title}{The quaternion model of doubly-fed induction motor}.
\newblock \bibinfo{journal}{In Proceedings of International Forum on Strategic Technology} , \bibinfo{pages}{32--36}.
\bibitem[{Pei and Cheng(1999)}]{Image-2}
\bibinfo{author}{Pei, S.C.}, \bibinfo{author}{Cheng, C.M.}, \bibinfo{year}{1999}.
\newblock \bibinfo{title}{Color image processing by using binary quaternion-moment-preserving thresholding technique}.
\newblock \bibinfo{journal}{IEEE Transactions on Image Processing} \bibinfo{volume}{8}, \bibinfo{pages}{614--628}.
\bibitem[{Phillips(2003)}]{Quantum-Physics}
\bibinfo{author}{Phillips, A.C.}, \bibinfo{year}{2003}.
\newblock \bibinfo{title}{Introduction to quantum mechanics}.
\newblock \bibinfo{publisher}{Wielly}.
\bibitem[{P\"{o}ppelbaum and Schwung(2022)}]{QBA}
\bibinfo{author}{P\"{o}ppelbaum, J.}, \bibinfo{author}{Schwung, A.}, \bibinfo{year}{2022}.
\newblock \bibinfo{title}{Quaternion backpropagation}.
\newblock \bibinfo{journal}{arXiv:2212.13082} .
\bibitem[{Said et~al.(2008)Said, Bihan and Sangwine}]{QFFT}
\bibinfo{author}{Said, S.}, \bibinfo{author}{Bihan, N.L.}, \bibinfo{author}{Sangwine, S.J.}, \bibinfo{year}{2008}.
\newblock \bibinfo{title}{Fast complexified quaternion \text{Fourier} transform}.
\newblock \bibinfo{journal}{IEEE Transactions on Signal Processing} \bibinfo{volume}{56}, \bibinfo{pages}{1522--1531}.
\bibitem[{Sangwine and Bihan(2005)}]{sangwine05}
\bibinfo{author}{Sangwine, S.}, \bibinfo{author}{Bihan, N.L.}, \bibinfo{year}{2005}.
\newblock \bibinfo{title}{{Quaternion Toolbox for Matlab [Online]}}.
\newblock \bibinfo{journal}{Available: http://qtfm.sourceforge.net/} .
\bibitem[{Seberry et~al.(2008)Seberry, Finlayson, Adams, Wysocki, Xia and Wysocki}]{Comm2}
\bibinfo{author}{Seberry, J.}, \bibinfo{author}{Finlayson, K.}, \bibinfo{author}{Adams, S.S.}, \bibinfo{author}{Wysocki, T.A.}, \bibinfo{author}{Xia, T.}, \bibinfo{author}{Wysocki, B.J.}, \bibinfo{year}{2008}.
\newblock \bibinfo{title}{The theory of quaternion orthogonal designs}.
\newblock \bibinfo{journal}{IEEE Transactions on Sognal Processing} \bibinfo{volume}{56}, \bibinfo{pages}{256--265}.
\bibitem[{Shoemake(1985)}]{CG}
\bibinfo{author}{Shoemake, K.}, \bibinfo{year}{1985}.
\newblock \bibinfo{title}{Animating rotation with quaternion curves}.
\newblock \bibinfo{journal}{ACM SIGGRAPH Computer Graphics} \bibinfo{volume}{19}, \bibinfo{pages}{245--254}.
\bibitem[{Spring(1986)}]{QREV}
\bibinfo{author}{Spring, K.W.}, \bibinfo{year}{1986}.
\newblock \bibinfo{title}{{Euler} parameters and the use of quaternion algebra in the manipulation of finite rotations: {A} review}.
\newblock \bibinfo{journal}{Mechanism and Machine Theory} \bibinfo{volume}{21}, \bibinfo{pages}{365--373}.
\bibitem[{Stern and Fischer(2018)}]{Comm1}
\bibinfo{author}{Stern, S.}, \bibinfo{author}{Fischer, R.F.H.}, \bibinfo{year}{2018}.
\newblock \bibinfo{title}{Quaternion-valued multi-user {MIMO} transmission via dual-polarized antennas and {QLLL} reduction}.
\newblock \bibinfo{journal}{In Proceedings of International Conference on Telecommunications} , \bibinfo{pages}{63--69}.
\bibitem[{Talebi(2016)}]{PouriaPhD}
\bibinfo{author}{Talebi, S.P.}, \bibinfo{year}{2016}.
\newblock \bibinfo{title}{Adaptive filtering algorithms for quaternion-valued signals}.
\newblock Ph.D. thesis. Imperial College London.
\bibitem[{Talebi et~al.(2025)Talebi, Cheong~Took and Mandic}]{QML}
\bibinfo{author}{Talebi, S.P.}, \bibinfo{author}{Cheong~Took, C.}, \bibinfo{author}{Mandic, D.P.}, \bibinfo{year}{2025}.
\newblock \bibinfo{title}{A quantum of learning: Using quaternion algebra to model learning on quantum devices}.
\newblock \bibinfo{journal}{In Proceedigns of International Conference on Digital Signal Processing} , \bibinfo{pages}{1--5}\DOIprefix\doi{10.1109/DSP65409.2025.11075094}.
\bibitem[{Talebi and Mandic(2015)}]{3PhaseMe}
\bibinfo{author}{Talebi, S.P.}, \bibinfo{author}{Mandic, D.P.}, \bibinfo{year}{2015}.
\newblock \bibinfo{title}{A quaternion frequency estimator for three-phase power systems}.
\bibitem[{\text{A. B. Carlson, P. B. Crilly, and J. C. Rutledge}(2002)}]{Comms}
\bibinfo{author}{\text{A. B. Carlson, P. B. Crilly, and J. C. Rutledge}}, \bibinfo{year}{2002}.
\newblock \bibinfo{title}{Communication systems: {An} Introduction to signals and noise in electrical communications}.
\newblock \bibinfo{edition}{4} ed., \bibinfo{publisher}{McGraw-Hill}.
\bibitem[{\text{C. C. Took and D. P. Mandic}(2010)}]{HR-Clive}
\bibinfo{author}{\text{C. C. Took and D. P. Mandic}}, \bibinfo{year}{2010}.
\newblock \bibinfo{title}{A quaternion widely linear adaptive filter}.
\newblock \bibinfo{journal}{IEEE Transactions on Signal Processing} \bibinfo{volume}{58}, \bibinfo{pages}{4427--4431}.
\bibitem[{\text{D. P. Mandic, C. Jahanchahi, and C. C. Took}(2011)}]{HR-Gradient}
\bibinfo{author}{\text{D. P. Mandic, C. Jahanchahi, and C. C. Took}}, \bibinfo{year}{2011}.
\newblock \bibinfo{title}{A quaternion gradient operator and its applications}.
\newblock \bibinfo{journal}{IEEE Signal Processing Letters} \bibinfo{volume}{18}, \bibinfo{pages}{47--50}.
\bibitem[{\text{P. J. Schreier and L. L. Scharf}(2003)}]{SOACRV}
\bibinfo{author}{\text{P. J. Schreier and L. L. Scharf}}, \bibinfo{year}{2003}.
\newblock \bibinfo{title}{Second-order analysis of improper complex random vectors and processes}.
\newblock \bibinfo{journal}{IEEE Transactions on Signal Processing} \bibinfo{volume}{51}, \bibinfo{pages}{714--725}.
\bibitem[{\text{S. Miron, N. Le Bihan, and J. I. Mars}(2006)}]{MUSIC}
\bibinfo{author}{\text{S. Miron, N. Le Bihan, and J. I. Mars}}, \bibinfo{year}{2006}.
\newblock \bibinfo{title}{Quaternion-\text{MUSIC} for vector-sensor array processing}.
\newblock \bibinfo{journal}{IEEE Transactions on Signal Processing} \bibinfo{volume}{54}, \bibinfo{pages}{1218--1229}.
\bibitem[{\text{T. Adali, P. J. Schreier, and L. L. Scharf}(2011)}]{CI}
\bibinfo{author}{\text{T. Adali, P. J. Schreier, and L. L. Scharf}}, \bibinfo{year}{2011}.
\newblock \bibinfo{title}{Complex-valued signal processing: The proper way to deal with impropriety}.
\newblock \bibinfo{journal}{IEEE Transactions on Signal Processing} \bibinfo{volume}{59}, \bibinfo{pages}{5101--5125}.
\bibitem[{Tobar and Mandic(2014)}]{QRKHS}
\bibinfo{author}{Tobar, F.A.}, \bibinfo{author}{Mandic, D.P.}, \bibinfo{year}{2014}.
\newblock \bibinfo{title}{Quaternion reproducing kernel {Hilbert} spaces: {Existence} and uniqueness conditions}.
\newblock \bibinfo{journal}{IEEE Transactions on Information Theory} \bibinfo{volume}{60}, \bibinfo{pages}{5736--5749}.
\bibitem[{Took and Mandic(2011)}]{AQS}
\bibinfo{author}{Took, C.C.}, \bibinfo{author}{Mandic, D.P.}, \bibinfo{year}{2011}.
\newblock \bibinfo{title}{Augmented second-order statistics of quaternion random signals}.
\newblock \bibinfo{journal}{Signal Processing} \bibinfo{volume}{91}, \bibinfo{pages}{214--224}.
\bibitem[{Ujang et~al.(2011)Ujang, Took and Mandic}]{QNAF}
\bibinfo{author}{Ujang, B.C.}, \bibinfo{author}{Took, C.C.}, \bibinfo{author}{Mandic, D.P.}, \bibinfo{year}{2011}.
\newblock \bibinfo{title}{Quaternion-valued nonlinear adaptive filtering}.
\newblock \bibinfo{journal}{IEEE Transactions on Neural Networks} \bibinfo{volume}{22}, \bibinfo{pages}{1193--1206}.
\newblock \DOIprefix\doi{10.1109/TNN.2011.2157358}.
\bibitem[{Voight(2021)}]{QuatBook}
\bibinfo{author}{Voight, J.}, \bibinfo{year}{2021}.
\newblock \bibinfo{title}{Quaternion Algebras}.
\newblock \bibinfo{publisher}{Springer}.
\bibitem[{Widrow et~al.(1975)Widrow, {McCool} and Ball}]{CLMS}
\bibinfo{author}{Widrow, B.}, \bibinfo{author}{{McCool}, J.}, \bibinfo{author}{Ball, M.}, \bibinfo{year}{1975}.
\newblock \bibinfo{title}{The complex \text{LMS} algorithm}.
\newblock \bibinfo{journal}{Proceedings of the IEEE} \bibinfo{volume}{63}, \bibinfo{pages}{719--720}.
\bibitem[{Xu et~al.(2015)Xu, Jahanchahi, Took and Mandic}]{GenHR}
\bibinfo{author}{Xu, D.}, \bibinfo{author}{Jahanchahi, C.}, \bibinfo{author}{Took, C.C.}, \bibinfo{author}{Mandic, D.P.}, \bibinfo{year}{2015}.
\newblock \bibinfo{title}{Enabling quaternion derivatives: {The} generalized {HR} calculus}.
\newblock \bibinfo{journal}{Royal Society Open Science} \bibinfo{volume}{2}.
\bibitem[{Zhang et~al.(2025)Zhang, Xiang, Zheng, Talebi and Mandic}]{CommME}
\bibinfo{author}{Zhang, M.}, \bibinfo{author}{Xiang, M.}, \bibinfo{author}{Zheng, Z.}, \bibinfo{author}{Talebi, S.P.}, \bibinfo{author}{Mandic, D.P.}, \bibinfo{year}{2025}.
\newblock \bibinfo{title}{A class of widely linear quaternion blind equalisation algorithms}.
\newblock \bibinfo{journal}{Signal Processing} \bibinfo{volume}{230}, \bibinfo{pages}{109863}.

\end{thebibliography}
%
%
%
%
%
%
%
%
%
%
%
%
%
%
%
%
%
\end{document}